\documentclass[10pt,twocolumn,letterpaper]{article}

\usepackage{cvpr}              

\usepackage{amsmath,amsfonts}
\usepackage{algorithmic}
\usepackage{algorithm}
\usepackage{array}
\usepackage{textcomp}
\usepackage{stfloats}
\usepackage{float}
\usepackage{url}
\usepackage{verbatim}
\usepackage{graphicx}
\usepackage{cite}
\usepackage{booktabs}
\usepackage{bm}
\usepackage{pifont}
\usepackage{enumitem}
\usepackage{soul}
\usepackage{makecell}
\usepackage{tabularx}
\usepackage{multirow}
\usepackage[dvipsnames,table]{xcolor}
\usepackage{siunitx}
\usepackage{balance}
\usepackage{lscape}
\usepackage[utf8]{inputenc}
\usepackage[T1]{fontenc}
\usepackage{pmboxdraw} 
\usepackage{tocloft}
\definecolor{cvprblue}{rgb}{0.21,0.49,0.74}
\usepackage[pagebackref,breaklinks,colorlinks,citecolor=cvprblue]{hyperref}

\usepackage{listings}
\definecolor{vscode-light-background}{RGB}{240,239,239}  
\definecolor{vscode-foreground}{RGB}{0,0,0}               
\definecolor{vscode-blue}{RGB}{56,92,170}                 
\definecolor{vscode-green}{RGB}{68,191,189}                
\definecolor{vscode-red}{RGB}{237,90,38}                  
\definecolor{vscode-purple}{RGB}{134,0,179}                
\definecolor{vscode-orange}{RGB}{214,103,0}                

\usepackage[capitalize]{cleveref}
\crefname{section}{Sec.}{Secs.}
\Crefname{section}{Section}{Sections}
\Crefname{table}{Table}{Tables}
\crefname{table}{Tab.}{Tabs.}

\def \ie {\emph{i.e.},}
\def \eg {\emph{e.g.},}

\def \wrt {\emph{w.r.t.}}

\newcolumntype{Y}{>{\centering\arraybackslash}X}

\newcommand{\tit}[1]{\smallbreak\noindent\textbf{#1.}}

\newcommand\blfootnote[1]{%
  \begingroup
  \renewcommand\thefootnote{}\footnote{#1}%
  \addtocounter{footnote}{-1}%
  \endgroup
}

\newcommand{\nocontentsline}[3]{}
\let\origcontentsline\addcontentsline
\newcommand\stoptoc{\let\addcontentsline\nocontentsline}
\newcommand\resumetoc{\let\addcontentsline\origcontentsline}

\def\paperID{13471} 
\def\confName{CVPR}
\def\confYear{2025}

\title{Zero-Shot Styled Text Image Generation, but Make It Autoregressive}

\author{
Vittorio Pippi$^{1^*}$ \quad Fabio Quattrini$^{1^*}$ \quad Silvia Cascianelli$^1$ \quad Alessio Tonioni$^{2}$ \quad Rita Cucchiara$^1$\\
\begin{tabular}{ccc}
\makecell{$^1$University of Modena and Reggio Emilia} & \makecell{$^2$Google}\\
\end{tabular}
}

\begin{document}
\maketitle
\stoptoc
\begin{abstract}
Styled Handwritten Text Generation (HTG) has recently received attention from the computer vision and document analysis communities, which have developed several solutions, either GAN- or diffusion-based, that achieved promising results. Nonetheless, these strategies fail to generalize to novel styles and have technical constraints, particularly in terms of maximum output length and training efficiency. To overcome these limitations, in this work, we propose a novel framework for text image generation, dubbed \textbf{Emuru}. Our approach leverages a powerful text image representation model (a variational autoencoder) combined with an autoregressive Transformer. Our approach enables the generation of styled text images conditioned on textual content and style examples, such as specific fonts or handwriting styles. We train our model solely on a diverse, synthetic dataset of English text rendered in over 100,000 typewritten and calligraphy fonts, which gives it the capability to reproduce unseen styles (both fonts and users' handwriting) in zero-shot. To the best of our knowledge, Emuru is the first autoregressive model for HTG, and the first designed specifically for generalization to novel styles. Moreover, our model generates images without background artifacts, which are easier to use for downstream applications. Extensive evaluation on both typewritten and handwritten, any-length text image generation scenarios demonstrates the effectiveness of our approach.\vspace{-0.3cm} 
\blfootnote{$^*$Equal contribution.}
\end{abstract}
    
\section{Introduction}
\label{sec:intro}
\begin{figure}
    \centering
    \includegraphics[width=\linewidth]{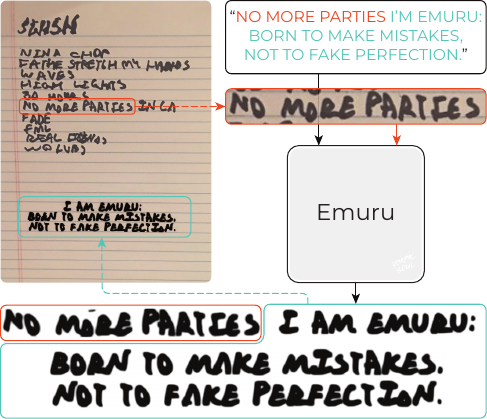}
    \caption{Our proposed Emuru model can generate images of any-length text lines mimicking out-of-distribution handwriting styles.}
    \label{fig:overview}\vspace{-0.5cm}
\end{figure}

The Styled Handwritten Text Generation (HTG) task entails generating images containing user-specified text rendered in a handwriting style resembling one or more reference style samples. 
In recent years, this task has regained significant attention from the Document Analysis and Computer Vision communities, especially due to its potential applications, ranging from generating synthetic data for other Document Analysis tasks to reproducing written styles to be used in graphic design and assistive technologies~\cite{haines2016my, bhunia2021handwriting, pippi2023choose}. 
From a technical point of view, handwriting can be treated as a sequence of strokes, as in the Online HTG paradigm, or as a static image, as in the Offline HTG paradigm. Note that Online HTG entails giving the model a style input in the form of strokes trajectories, and letting the model generate strokes trajectories to be rendered on an image. Therefore, Online HTG models are implemented as sequential models~\cite{graves2013generating, aksan2018deepwriting,aksan2018stcn,dai2023disentangling,luhman2020diffusion,ren2023diff}. Note that collecting data to train and use Online HTG models involves using digital pens and pads, which is cumbersome and costly and cannot be applied in scenarios where the style to reproduce is from an existing manuscript, \eg~historical documents. For these reasons, Offline HTG, where the style input and the generated output are both static images, has gained more popularity and is also the focus of this work. 

Despite the increasing research efforts in Offline HTG, generating synthetic images of handwritten text in a writer-specific style is still challenging. Specifically, current State-of-the-Art models: A. Struggle to generalize to handwriting styles that are significantly different from those encountered in training;
B. Are implemented by making some dataset-specific assumptions due to technical limitations of the architecture adopted, \eg~normalizing character width or enforcing a maximum text line length;  
C. Cannot disentangle properly the handwriting style (\ie~the shape of the strokes) from the reference image background, and thus, generate undesired artifacts in the output images.
To overcome these limitations, in this work, we take a step towards generalizable few-shot styled text image generation and propose Emuru (\Cref{fig:overview}): a continuous-token autoregressive image generation model that can generate text images of any length (B.) with high style fidelity and readability, while increasing generalization to novel styles (A.) and reducing background artifacts (C.). To the best of our knowledge, Emuru is the first autoregressive Offline styled HTG model proposed in the literature.

Emuru consists of a VAE that maps the style sample images in an embedding space where they become a varying-length sequence of continuous vectors, and an autoregressive Transformer Encoder-Decoder that takes as input the desired style embeddings sequence, the text contained in the style image, and the desired text, and iteratively generates an image containing the desired text in the desired style.
The generation is implemented as an autoregressive loop that outputs visual embeddings, which are then fed to the VAE Decoder to obtain the final output image. Given the autoregressive nature of the model, we train it to automatically establish when to stop the generation, thus removing any technical constraint on the maximum output length obtainable.
Both the VAE and the autoregressive Transformer in Emuru are trained (separately, in two stages) on a specifically-designed, purely synthetic dataset\footnote{\scriptsize{\url{https://huggingface.co/blowing-up-groundhogs/font-square-v2}}}. 
The images in the synthetic dataset are obtained by rendering a large amount of different English text lines (whose characters frequency is designed to be almost uniform) on different background images and with over 100k different fonts, both handwritten (calligraphy) and typewritten. Note that by working with a synthetic dataset, we have all the necessary information for training the VAE to reconstruct the text line image without the background. This way, the learned embeddings encode only the writing style, which gives Emuru the capacity to generate images without background artifacts. Moreover, training style variety and textual character frequency uniformity make Emuru able to generalize better to unseen styles, both typewritten and handwritten, and languages  (\ie~characters arrangements).

To validate our proposed approach, we perform an extensive experimental analysis that considers multiple real datasets of handwritten text in multiple Latin alphabet languages. Moreover, we introduce a novel dataset of typewritten and handwritten text images to foster the development of HTG models able to generalize also to typewritten styles. The results of our analysis demonstrate the efficacy of the proposed architectural and training solutions for generating single words and entire lines with increased generalization capability to new styles, both handwritten and typewritten with respect to the State-of-the-Art. The code and weights for Emuru are publicly available\footnote{\scriptsize{\url{https://huggingface.co/blowing-up-groundhogs/emuru}}}.

\section{Related Work}
\label{sec:related}

The early-proposed approaches for HTG~\cite{wang2005combining, lin2007style, thomas2009synthetic, haines2016my} entailed manually segmenting text images into glyphs, which were later recombined based on their handcrafted geometric features. Such approaches did not offer flexibility in terms of styles or text and required intensive human intervention. In recent years, the HTG task has re-gained popularity thanks to the capabilities of learning-based solutions, especially Offline HTG ones. 

\tit{Offline HTG}
Offline HTG approaches generate images of text containing a user-specified content string. Existing works implementing this paradigm rely either on GAN-based or Diffusion Model-based solutions. 
Early learning-based offline HTG works~\cite{alonso2019adversarial,fogel2020scrabblegan} were only able to condition on the desired content but did not offer any control over the handwriting style. This capability was later introduced in~\cite{kang2020ganwriting}, which proposed to condition the generation also on a representation of the style. In this respect, the style reference can consist either of a writer identifier~\cite{luo2022slogan,zhu2023conditional,nikolaidou2023wordstylist} (which completely prevents the models from imitating unseen styles) or some reference images in the form of: a paragraph~\cite{mayr2024zero}, a line~\cite{davis2020text}, fifteen~\cite{kang2020ganwriting, bhunia2021handwriting, pippi2023handwritten, vanherle2024vatr++} to five~\cite{nikolaidou2024diffusionpen} words, or a single word~\cite{gan2021higan, gan2022higan+, krishnan2023textstylebrush, dai2024one}. We argue that using more style reference information (\ie~more than one word image) can lead to superior performance and that it is still easy for a user to provide it. For this reason, in this work, we take a single image of a text line as style input.

Most HTG models focus on single words, \ie~short sequences of text tokens. As a result, when these models are used to generate long sequences, they struggle to render all the characters with the same quality and coherent proportions. Moreover, the words rendered are generated inside fixed-height canvases. This can result in different scales and misaligned baselines of words depending on the fact that they contain or do not contain characters with ascenders and descenders. For this reason, using existing HTG models to obtain long pieces of text is cumbersome and cannot be satisfactorily done by simply stitching images of words. To overcome this limitation, some works have been specifically trained to handle text lines~\cite{davis2020text,kang2021content} and paragraphs~\cite{mayr2024zero}. As an alternative applicable to output length-constrained HTG models (such as Diffusion Model-based approaches), the authors in~\cite{nikolaidou2024diffusionpen} have proposed a specific algorithm to stitch and blend short images to obtain longer text. In this work, to enforce proportionality between characters and word baseline alignment, we propose to exploit an autoregressive generative model without constraints on the output size. In particular, we use a single line as the style reference and ask our model to generate entire text lines, automatically establishing when to stop.
\begin{figure*}[t!]
    \centering
    \includegraphics[width=\linewidth]{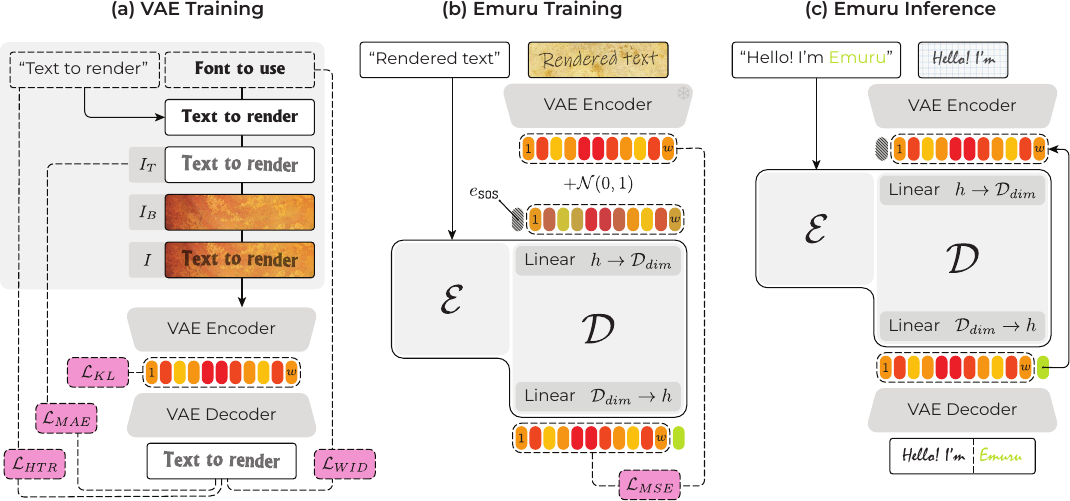}
    \caption{Our Emuru approach consists of a VAE (a) and an autoregressive Transformer Encoder-Decoder (b), both trained on a massive synthetic dataset of texts rendered in different fonts. At inference time (c), Emuru is given a reference style image, the text in the reference style image, and the desired text and is tasked to iteratively generate the output styled image, autonomously deciding when to stop.}
    \vspace{-0.3cm} 
    \label{fig:pipeline}
\end{figure*}
\vspace{-.3em}
\tit{Autoregressive image generation}
Following the success of Transformer-based natural text generation solutions~\cite{vaswani2017attention,radford2019language},
autoregressive image generation has been explored extensively. In most cases, image tokenizers are used to convert continuous images into discrete tokens, and generation is performed as next-token prediction~\cite{van2017neural,razavi2019generating,esser2021taming, ramesh2021zero,yu2022scaling}. However, State-of-the-Art performance in image generation is currently mostly obtained by using Diffusion Models~\cite{sohl2015deep,ho2020denoising,dhariwal2021diffusion,rombach2022high}, which have proven themselves effective in modeling continuous data. Recently, LlamaGen~\cite{sun2024autoregressive} and GIVT~\cite{tschannen2025givt} have shown the potential of continuous autoregressive generative models. In particular, GIVT~\cite{tschannen2025givt} leverages a continuous-latent $\beta$-VAE~\cite{higgins2017beta} and a Transformer Decoder to obtain superior performance compared to quantized VAEs. Removing the vector quantization and using continuous latents allows avoiding excessive information compression and the optimization problems in training~\cite{van2017neural,chang2022maskgit,huh2023straightening,mentzer2023finite}. Nonetheless, generative methods for natural images differ from styled HTG ones because they often assume a fixed number of output tokens, either because the image size is fixed or because it is one of the user inputs. On the other hand, for HTG, the length of the generated styled text image depends on the input style and content. To address this aspect, we propose to let the model decide when to stop generating, similar to what is done in natural text generation~\cite{vaswani2017attention,radford2019language}.

\section{Method}
\label{sec:method}

Our framework for text-image generation consists of a powerful text-image representation model (a VAE) combined with an autoregressive Transformer Encoder-Decoder to generate text images conditioned on the desired textual content and a style example, both of which can be of any length. 
Specifically, the VAE maps the styled text images into a dense latent space and reconstructs them without background. The Transformer autoregressively generates embeddings compatible with those in the VAE latent space, conditioned on the embeddings of the reference style image and the desired text. Then, the VAE Decoder can be used to obtain the final image.
A schematic representation of our approach is depicted in~\Cref{fig:pipeline}.

\subsection{Emuru Training Data}
State-of-the-Art HTG models are usually trained on a single dataset at a time, which is quite limiting due to the small variety of featured styles and text in the training images. This causes many HTG models to have a performance drop when generating images that are out-of-distribution for the style and the textual content, \eg~from a different dataset. 
In light of these considerations and inspired by the effectiveness of the synthetic pretraining of the style Encoder in~\cite{pippi2023handwritten, pippi2023evaluating}, we develop a massive synthetic dataset, which we use to train our HTG model. The dataset features RGB images containing text lines written over different backgrounds. 
As for the text content, we collect words from multiple English corpora available on the NLTK library\footnote{\scriptsize{\url{https://www.nltk.org/}}} and combine 1 to 32 of them into text lines $T$, making sure that both rare and common words are both well represented.
Moreover, we collect over 100k fonts available online and render the text lines using these fonts on greyscale images, where the background is always white. Finally, we apply several random geometric transformations to the images (see the Supplementary for details). This way, we obtain the text image $I_T$ for each sample in the synthetic dataset. Then, we superimpose $I_T$ to an RGB image $I_B$ depicting a plausible background for text (\eg~paper), randomly selected among a pre-defined set. This way, we obtain the synthetic sample images $I$ with shape $W \times H$ where $H=64$ and $W$ scales proportionally to the length of the text. In total, we obtain 2.2 million text images. Further details about the dataset are given in the Supplementary.
Note that we solely use this synthetic dataset for training both the components of our proposed approach: first, the VAE and, in a later stage, the autoregressive Transformer, keeping the VAE frozen.

\subsection{Emuru VAE}
First, we train a $\beta$-VAE to encode the text images into a compressed latent space. We choose a convolutional architecture similar to those commonly used in Stable Diffusion Models~\cite{rombach2022high}. The VAE Encoder takes as input the images $I^{3 \times W \times H}$ and maps them in a $c \times h \times w$ embedding tensor. Due to the nature of lines of text, \ie~arbitrary length but mostly fixed height, we model the images as a $w$-long sequence of $h \times c$ vectors. Intuitively, each of the $w$ vectors encodes a vertical slice of the text line. The Decoder is fed with these vectors and is tasked to reconstruct the grayscale text image $I_T$. 

We train the VAE by using a Reconstruction loss expressed in terms of $L_1$ distance between the ground truth $I_T$ image (\ie~only the text without background) and the reconstructed output ($\mathcal{L}_{MAE}$) and the Kullback–Leibler divergence loss ($\mathcal{L}_{KL}$). In line with \cite{rombach2022high,tschannen2025givt}, we weigh this latter loss term with $\beta << 1$ in order to prioritize reconstruction quality rather than constraining a strongly regularized latent space~\cite{higgins2017beta}.
Additionally, we incorporate supervision signals (namely, a Cross-Entropy loss $\mathcal{L}_{WID}$ and a CTC loss $\mathcal{L}_{HTR}$) from two auxiliary networks. The first is a writing style classification network, pretrained on our synthetic dataset to identify the 100k styles from the reconstructed image, which enforces style fidelity. The second is a Handwritten Text Recognition (HTR) model, also pretrained on the synthetic dataset, tasked to recognize the text $T$ in the reconstructed image to enforce readability.

\subsection{Emuru Autoregressive Transformer}
Once we obtained a strong latent image representation with the VAE, we train a generative model by using an autoregressive Transformer featuring a Transformer Encoder $\mathcal{E}$ and a Transformer Decoder $\mathcal{D}$, architecturally similar to T5~\cite{raffel2020exploring}. Both are trained on samples from the devised synthetic dataset.
Specifically, at training time, $\mathcal{E}$ takes as input a sequence of 
characters in the desired content string ($T_{out}$) tokenized by using the single-character tokenizer. Then, self-attention is computed among these tokens to obtain a sequence of text embeddings $\mathbf{s} = [s_1, ..., s_{k}]$. On the other hand, the input to the Transformer Decoder $\mathcal{D}$ is obtained as follows. First, we feed the pretrained, frozen VAE Encoder with the style image $I$ to obtain a $w$-long sequence of $h$-dimensional embeddings, $\mathbf{v} = [v_1, ..., v_{w}]$. To reduce the risk of exposure bias at inference time, these embeddings are summed to some noise sampled from a Normal distribution. 
Then, the resulting noisy embeddings are linearly projected into $\mathcal{D}_{dim}$-dimensional vectors, to which we prepend a special start-of-sequence ($\mathtt{SOS}$) learned embedding. The resulting $\mathbf{e} = [e_{\mathtt{SOS}}, e_1, ..., e_{w}]$ vectors are then fed to $\mathcal{D}$, which autoregressively predicts a sequence of vectors $\mathbf{e'} = [e'_1, ..., e'_{w}]$.
To this end, causal masked self-attention is performed on the $\mathbf{e}$'s, and unmasked cross-attention is performed between the $\mathbf{e}$'s and the $\mathbf{s}$'s from $\mathcal{E}$.  
The so-obtained $\mathbf{e'}$ sequence is projected back into the $\mathbf{v'}$ sequence of $h$-dimensional vectors that we use to compute the training loss. 

We train the Transformer following a teacher-forcing strategy and using the MSE loss between the predicted $\mathbf{v}'$ and ground truth $\mathbf{v}$ embeddings ($\mathcal{L}_{MSE}$). Moreover, to make Emuru handle sequences of different lengths, we take inspiration from the curriculum learning strategy adopted in~\cite{kang2021content}. The authors proposed to train their HTG model in three stages, with non-overlapping sets of increasingly longer images. Here, we train Emuru in two stages and increase the variability of the training images maximum length. Specifically, we first train on text lines containing from 4 to 7 words, then we fine-tune on lines containing from 1 to 32 words. 
Note that using only a single loss term in the form of the MSE makes the training process more controllable, thanks to the limited number of hyperparameters involved, and less resource-demanding compared to adversarial and diffusion process-based approaches that also rely on auxiliary networks to obtain regularization signals. 

\subsection{Inference-time behavior}
\begin{figure}
    \centering
    \includegraphics[width=0.85\linewidth]{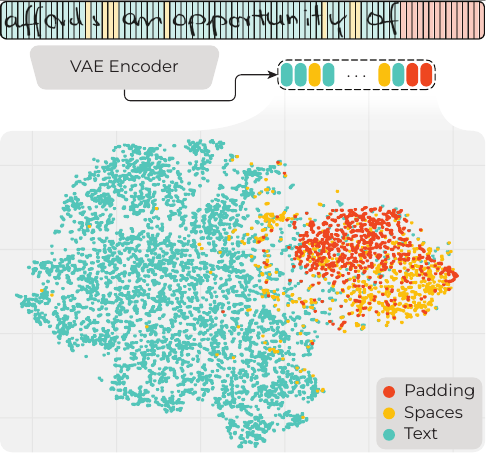}
    \caption{Characters and padding VAE embeddings distribution obtained via t-SNE.}
    \label{fig:eos}\vspace{-0.3cm}
\end{figure}

At inference time, we feed Emuru with a reference style image containing a text line ($I_{style}$) and the concatenation of two strings: one with the text in the reference style image ($T_{style}$), and one with the desired output text ($T_{out}$). Emuru is tasked to continue the generation from $I_{style}$ by following the desired text $T_{out}$, in the style of the reference image. 
Specifically, at each generation step $t$, $\mathcal{D}$ predicts the $(w+1)$-th embedding ($e_{w+1}$) based on the embeddings of $I_{style}$ and the $t$ previously-generated embeddings. 
Note that all training images used for our model have some padding in the form of white spaces since it is required for batched execution. Therefore, a training sample for Emuru is consistently modeled as a sequence of embeddings relative to characters followed by embeddings relative to white padding. Looking at the VAE latent space, it can be observed that the latter are quite distinguishable from the text characters embeddings (see~\Cref{fig:eos}). Thus, we stop the iterative generation of embeddings when Emuru emits $P$ consecutive padding embeddings, which act as an implicit end-of-sequence signal. $P$ is set to $10$ after an empirical analysis to minimize the risk of early stopping the generation. 
At the end of the iterative generation, the new sequence of the generated $h$-dimensional vectors (excluding the last $P$ padding vectors) is fed to the pretrained, frozen VAE Decoder to obtain the final image.

\section{Experiments}
\label{sec:experiments}
Despite the recently increased interest in HTG, the proposed works on the task do not follow a common, standard evaluation protocol. To fairly assess our contribution against the State-of-the-Art, we re-run the previous approaches for standardizing the evaluation setup. In the following, we give the details of the evaluation protocol we follow and of the competitors considered, which are models for which code and weights were made publicly available. We hope this setting to be embraced by the community as a new standardized benchmark. 

\tit{Considered Scores}
To evaluate the performance of our approach by considering different aspects of the HTG task, we adopt multiple scores in our analysis. These are the Fréchet Inception Distance (\textbf{FID})~\cite{heusel2017gans} and the Kernel Inception Distance (\textbf{KID})~\cite{binkowski2018demystifying}, commonly adopted for image generation, and the task-specific Handwriting Distance (\textbf{HWD})~\cite{pippi2023hwd}. Moreover, we introduce two other scores. The first is the Binarized FID (\textbf{BFID}), obtained by computing the FID on binarized reference and generated images to capture mostly the perceptual handwriting style and disregard color or textures. The second is the Absolute Character Error Rate Difference (\textbf{$\Delta$CER}). To compute it, we consider the State-of-the-Art text recognition TrOCR-Base~\cite{li2021trocr} model and compute its CER on the reference and generated images to capture their readability while taking into consideration the style. Intuitively, we want the generated images to be as readable as the reference ones but not significantly more, which could mean that the HTG model has collapsed to a very readable style, different from the reference one. Note that we also considered additional scores for our evaluation, which are reported in the Supplementary.

\tit{Considered Datasets}
Our proposed Emuru is trained only on a large synthetic dataset of images containing English text rendered in calligraphy and typewritten fonts. To assess its generalization capabilities, we consider multiple multi-writer datasets and apply Emuru without any training or fine-tuning. These datasets are the commonly adopted \textbf{IAM}~\cite{marti2002iam} dataset, both in its word-level and line-level format, and the line-level annotated \textbf{CVL}~\cite{kleber2013cvl} and \textbf{RIMES}~\cite{augustin2006rimes} datasets. Moreover, we devise a dataset of text lines images obtained by rendering song lyrics in multiple Latin-alphabet languages (English, French, German, and Italian) with 100 publicly available fonts\footnote{\scriptsize{\url{https://fonts.google.com/}}}, both calligraphy and typewritten, on a white background. To foster the research on styled text image generation strategies able to generalize also to typewritten styles, we will release this dataset, dubbed \textbf{Karaoke}. Further details about these datasets are reported in the Supplementary. Note that to ensure a consistent and fair evaluation, we define a fixed set of reference style images and desired text to guide the generation for Emuru and the considered competitors, described below.

\tit{Considered State-of-the-Art approaches}
In our comparison, we consider State-of-the-Art approaches for styled HTG following both the adversarial training paradigm and the denoising diffusion paradigm. Note that for consistent evaluation, we select models for which code and weights have been made available by the respective authors. Specifically, we include in our analysis the Convolutional GAN-based \textbf{HiGAN+}~\cite{gan2022higan+} and \textbf{TS-GAN}~\cite{davis2020text}, and the Transformer GAN-based \textbf{HWT}~\cite{bhunia2021handwriting}, \textbf{VATr}~\cite{pippi2023handwritten}, and \textbf{VATr++}~\cite{vanherle2024vatr++}. Moreover, we consider the recently-proposed \textbf{DiffPen}~\cite{nikolaidou2024diffusionpen} and \textbf{One-DM}~\cite{dai2024one} HTG Diffusion Models. 

\tit{Implementation Details}
We train all the components of our model on a single NVIDIA RTX 4090.

Emuru convolutional VAE has 4 layers for the Encoder and the Decoder, with the output channels for each block set to 32, 64, 128, and 256.
The Encoder spatially downsamples the input $I^{3{\times}W{\times}H}$ to a 1-channel latent vector $h{\times}w$, where $h{=}H/8$ and $w{=}W/8$. We use one channel for the latent space to compress the information on the styled images and allow a lightweight Transformer Decoder to effectively handle the resulting latents. We train the VAE for 60k iterations with the AdamW~\cite{loshchilov2018decoupled} optimizer and learning rate 1e-4. 
The loss is composed of four terms: $\mathcal{L}_{MAE}$, $\mathcal{L}_{WID}$, $\mathcal{L}_{HTR}$, and $\mathcal{L}_{KL}$, combined with weight 1, 0.005, 0.3, and $\beta=$1e-6, respectively. Note that for the HTR supervision, we employ noisy teacher-forcing with probability 0.3.
As for the auxiliary networks, the writer identification model is a ResNet with 6 blocks trained with the AdamW optimizer, learning rate 1e-4, and batch size 256 until it reached 60\% accuracy on the synthetic dataset. The HTR network is a Transformer Encoder-Decoder~\cite{kang2020pay} trained with AdamW, learning rate 1e-4, and batch size 128 until it reached 0.25 CER on the synthetic dataset. 

The base architecture of the autoregressive Transformer in Emuru is a T5-Large~\cite{raffel2020exploring} Encoder-Decoder with two linear adapters, one before the input of the Decoder and another that replaces the original T5 Decoder linear classifier. The adapter layers convert the embedding sequence from the VAE feature space $h=8$ to the T5-Decoder inner dimension $\mathcal{D}_{dim}=1024$ and vice versa. The text input of the Encoder is tokenized with the single-character tokenizer Google-ByT. 
For the first stage of training, we adopt the AdamW optimizer with learning rate 1e-4 and weight decay 1e-2 and train for 70k iterations with batch size 256, employing noisy teacher-forcing with probability 0.1. 
All images used in this phase contain from 4 to 7 words and are padded or cropped to a fixed width of 768px. For the second stage of training, we use images containing from 1 to 32 words, padded or cropped to a fixed width of 2048. For this phase, we train for 250 iterations with a virtual batch size of 256 with gradient accumulation, without employing noisy teacher-forcing.

\subsection{Results}
\begin{table}[t]
    \centering
    \setlength{\tabcolsep}{.5em}
    \resizebox{\columnwidth}{!}{%
    \begin{tabular}{l ccccc}
    \toprule
    & \textbf{FID$\downarrow$}
    & \textbf{BFID$\downarrow$}
    & \textbf{KID$\downarrow$}
    & \textbf{$\Delta$CER$\downarrow$}
    & \textbf{HWD$\downarrow$} \\
    \midrule
    \textbf{SD1.5 VAE}  & 29.39 & 7.36 & 32.14 & \textbf{0.00} & 0.77 \\   
    \textbf{SD3 VAE}    & 21.90 & 3.61 & 23.01 & \textbf{0.00} & \textbf{0.74} \\
    \textbf{Emuru VAE}  & \textbf{19.22} & \textbf{1.62} & \textbf{16.35} & 0.03 & 0.85 \\
    \bottomrule
    \end{tabular}
     }
    \caption{Comparison between the VAE featured in Emuru and other VAEs used in Diffusion Model-based HTG solutions when reproducing the images in the test set of different HTG datasets. We report the mean of each score obtained over all the considered datasets. The KID is multiplied by $10^3$ and the best performance is in bold for readability.}
    \label{tab:best_vae}\vspace{-0.3cm}
\end{table}

\tit{VAE Reconstruction Performance}
Since our autoregressive model generates only latent vectors, it is clear that the performance of Emuru is tied to the VAE reconstruction quality. Therefore, we start our experiments by evaluating the quality of the VAE, which is tasked to embed the styled text images into the latent space and perform background removal. We perform extensive analysis on the reconstruction task over multiple datasets and compare Emuru VAE to other VAEs pretrained for natural images from State-of-the-Art Diffusion Models, namely Stable Diffusion 1.5 (SD1.5)~\cite{rombach2022high} and Stable Diffusion 3 (SD3)~\cite{esser2024scaling}. We report the average scores over our considered datasets in~\Cref{tab:best_vae} and the dataset-specific results in the Supplementary. As it emerges from this analysis, our text images-tailored VAE achieves good generalization and style consistency compared to generalist VAEs, also enabling downstream lightweight token modeling using only $\sim$16\% of their parameters and having a single latent channel (note that the VAE has 4 channels in SD1.5 and 16 channels in SD3). 
\begin{table}[t]
    \centering
    \setlength{\tabcolsep}{.5em}
    \resizebox{\columnwidth}{!}{%
    \begin{tabular}{l ccccc}
    \toprule
    & \multicolumn{5}{c}{\textbf{IAM Words}}\\ 
    \cmidrule{2-6}
    & \textbf{FID$\downarrow$}
    & \textbf{BFID$\downarrow$}
    & \textbf{KID$\downarrow$}
    & \textbf{$\Delta$CER$\downarrow$}
    & \textbf{HWD$\downarrow$} \\
    \midrule
    \textbf{TS-GAN}          & 129.57 & 86.45 & 141.08 & 0.28 & 4.22 \\
    \textbf{HiGAN+}          & ~50.19 & 21.92 & ~43.39 & 0.20 & 3.12 \\
    \textbf{HWT}             & ~27.83 & 15.09 & ~19.64 & 0.15 & 2.01 \\
    \textbf{VATr}            & ~30.26 & 15.81 & ~22.31 & \textbf{0.00} & 2.19 \\
    \textbf{VATr++}          & ~31.91 & 17.15 & ~23.05 & 0.07 & 2.54 \\
    \textbf{One-DM}          & ~27.54 & 10.73 & ~21.39 & 0.10 & 2.28 \\
    \textbf{DiffPen}         & ~\textbf{15.54} & ~\textbf{6.06} & ~\textbf{11.55} & 0.06 & \textbf{1.78} \\
    \textbf{Emuru}           & ~63.61 & 37.73 & ~62.34 & 0.19 & 3.03 \\
    \midrule
        & \multicolumn{5}{c}{\textbf{IAM Lines}}\\ 
    \cmidrule{2-6}
    & \textbf{FID$\downarrow$}
    & \textbf{BFID$\downarrow$}
    & \textbf{KID$\downarrow$}
    & \textbf{$\Delta$CER$\downarrow$}
    & \textbf{HWD$\downarrow$}\\
    \midrule
    \textbf{TS-GAN}          & 44.17 & 19.45 & 45.42 & 0.02 & 3.21 \\
    \textbf{HiGAN+}          & 74.41 & 34.18 & 77.27 & \textbf{0.00} & 3.25 \\
    \textbf{HWT}             & 44.72 & 30.26 & 43.49 & 0.33 & 2.97 \\
    \textbf{VATr}            & 35.32 & 27.97 & 33.61 & 0.02 & 2.37 \\
    \textbf{VATr++}          & 34.00 & 21.67 & 29.68 & 0.03 & 2.38 \\
    \textbf{One-DM}          & 43.89 & 21.54 & 44.48 & 0.13 & 2.83 \\
    \textbf{DiffPen}         & \textbf{12.89} & ~6.87 & ~\textbf{9.73} & 0.03 & 2.13 \\
    \textbf{Emuru}           & 13.89 & ~\textbf{6.19} & 11.30 & 0.14 & \textbf{1.87} \\ 
    \bottomrule
    \end{tabular}
    }\caption{Comparison on the word-level and line-level IAM datasets between Emuru and State-of-the-Art approaches trained on IAM. The KID is multiplied by $10^3$ and the best performance is in bold for readability.}
    \label{tab:iam}\vspace{-0.3cm}
\end{table}
\begin{figure*}[t]
    \centering
    \includegraphics[width=\textwidth]{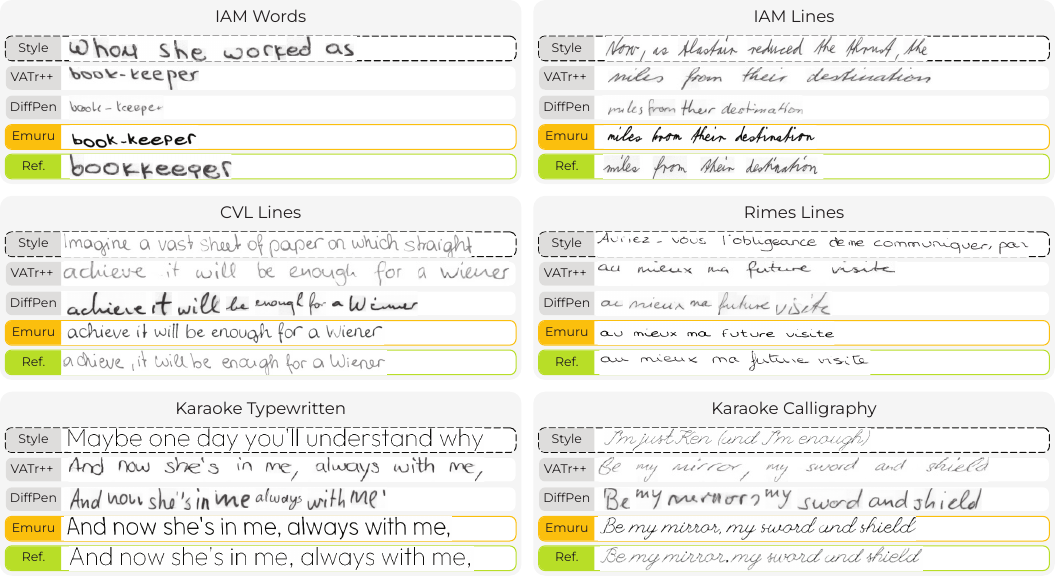}
    \caption{Qualitative comparison between Emuru, the GAN-based VATr++, and the Diffusion Model-based DiffPen when generating images from the considered datasets. We report the input style image used for guiding the generation and another reference image in the same style. We let the models generate the same text as in the reference to better observe the style imitation capabilities of the models.}
    \label{fig:qualitative_comparison}\vspace{-0.2cm}
\end{figure*}
\tit{Zero-Shot inference on IAM}
Current HTG models achieve State-of-the-Art performance by training on a single dataset at a time (generally, the word-level IAM). In~\Cref{tab:iam}, we compare IAM-trained competitors with Emuru, trained only on the proposed synthetic dataset. We refer to this setting as \textit{zero-shot inference} because the model has never seen IAM samples during training. The performance of our approach is mostly in line with that of existing methods, even achieving top scores for line generation in terms of BFID and HWD. As these scores focus on style rather than background, this indicates that the generated lines are more representative of the input style. Moreover, as shown in~\Cref{fig:qualitative_comparison}, scores such as FID and KID are not representative of aspects like word alignment~\cite{pippi2023hwd}, which is a weakness of models that generate single words and stitch them together to create a sentence, like \eg~DiffPen. Despite not having been trained on IAM samples, Emuru demonstrates strong generalization, performing on par with competitors trained on IAM.

\begin{table}[]
    \centering
    \setlength{\tabcolsep}{.5em}
    \resizebox{\columnwidth}{!}{%
    \begin{tabular}{l ccccc}
    \toprule
    & \multicolumn{5}{c}{\textbf{CVL Lines}}\\ 
    \cmidrule{2-6}
    & \textbf{FID$\downarrow$}
    & \textbf{BFID$\downarrow$}
    & \textbf{KID$\downarrow$}
    & \textbf{$\Delta$CER$\downarrow$}
    & \textbf{HWD$\downarrow$} \\
    \midrule
    \textbf{TS-GAN}          & 42.12 & 31.97 & 43.15 & 0.13 & 3.07 \\
    \textbf{HiGAN+}          & 78.44 & 39.47 & 80.39 & 0.12 & 3.07 \\
    \textbf{HWT}             & 31.22 & 16.73 & 26.14 & 0.38 & 2.59 \\
    \textbf{VATr}            & 34.40 & 24.64 & 32.21 & 0.06 & 2.36 \\
    \textbf{VATr++}          & 35.53 & 19.87 & 34.15 & 0.12 & 2.18 \\
    \textbf{One-DM}          & 60.45 & 26.58 & 64.13 & 0.06 & 2.66 \\
    \textbf{DiffPen}         & 40.40 & 17.50 & 38.21 & \textbf{0.01} & 2.99 \\
    \textbf{Emuru}           & \textbf{14.39} & \textbf{10.77} & \textbf{12.34} & 0.13 & \textbf{1.82} \\
    \midrule
        & \multicolumn{5}{c}{\textbf{RIMES Lines}}\\ 
    \cmidrule{2-6}
    & \textbf{FID$\downarrow$}
    & \textbf{BFID$\downarrow$}
    & \textbf{KID$\downarrow$}
    & \textbf{$\Delta$CER$\downarrow$}
    & \textbf{HWD$\downarrow$}\\
    \midrule
    \textbf{TS-GAN}          & 109.04 & 36.39 & 132.90 & 0.12 & 3.26 \\
    \textbf{HiGAN+}          & 160.57 & 47.38 & 183.82 & 0.14 & 3.39 \\
    \textbf{HWT}             & 118.21 & 35.26 & 128.66 & 0.45 & 3.36 \\
    \textbf{VATr}            & 113.76 & 30.21 & 114.21 & 0.07 & 3.09 \\
    \textbf{VATr++}          & 110.04 & 35.61 & 104.05 & 0.10 & 2.83 \\
    \textbf{One-DM}          & 121.18 & 36.07 & 121.67 & 0.20 & 3.36 \\
    \textbf{DiffPen}         & ~89.79 & 18.25 & ~94.78 & \textbf{0.04} & 2.58 \\
    \textbf{Emuru}           & ~\textbf{26.93} & \textbf{13.26} & ~\textbf{21.19} & 0.25 & \textbf{2.18} \\
    \bottomrule
    \end{tabular}
    }\caption{Comparison on the line-level CVL and RIMES datasets between Emuru and state-of-the-art approaches trained on IAM. The KID is multiplied by $10^3$ and the best performance is in bold for readability.}
    \label{tab:cvl_rimes}
\end{table}
\begin{table}[]
    \centering
    \setlength{\tabcolsep}{.5em}
    \resizebox{\columnwidth}{!}{%
    \begin{tabular}{l ccccc}
    \toprule
    & \multicolumn{5}{c}{\textbf{Karaoke Calligraphy}}\\ 
    \cmidrule{2-6}
    & \textbf{FID$\downarrow$}
    & \textbf{BFID$\downarrow$}
    & \textbf{KID$\downarrow$}
    & \textbf{$\Delta$CER$\downarrow$}
    & \textbf{HWD$\downarrow$} \\
    \midrule
    \textbf{TS-GAN}          & 60.30 & 12.68 & 64.80 & 0.23 & 4.59 \\
    \textbf{HiGAN+}          & 125.75 & 69.41 & 136.75 & 0.08 & 4.90 \\
    \textbf{HWT}             & 62.69 & 43.03 & 59.35 & 0.32 & 4.50 \\
    \textbf{VATr}            & 72.22 & 47.66 & 67.70 & 0.05 & 3.89 \\
    \textbf{VATr++}          & 67.16 & 46.53 & 58.57 & \textbf{0.01} & 3.96 \\
    \textbf{One-DM}          & 59.73 & 38.30 & 56.55 & 0.04 & 4.31 \\
    \textbf{DiffPen}         & 34.19 & 25.78 & 28.91 & 0.16 & 4.18 \\
    \textbf{Emuru}           & \textbf{13.87} & ~\textbf{7.99} & ~\textbf{9.24} & 0.13 & \textbf{2.24} \\
    \midrule
        & \multicolumn{5}{c}{\textbf{Karaoke Typewritten}}\\ 
    \cmidrule{2-6}
    & \textbf{FID$\downarrow$}
    & \textbf{BFID$\downarrow$}
    & \textbf{KID$\downarrow$}
    & \textbf{$\Delta$CER$\downarrow$}
    & \textbf{HWD$\downarrow$}\\
    \midrule
    \textbf{TS-GAN}          & 141.41 & 75.78 & 157.33 & 0.32 & 4.70 \\
    \textbf{HiGAN+}          & 135.34 & 63.39 & 146.34 & 0.07 & 5.19 \\
    \textbf{HWT}             & 72.78 & 37.40 & 62.77  & 0.37 & 4.57 \\
    \textbf{VATr}            & 80.38 & 41.02 & 70.46  & 0.05 & 4.14 \\
    \textbf{VATr++}          & 76.03 & 41.69 & 63.17 & \textbf{0.01} & 4.15 \\
    \textbf{One-DM}          & 70.75 & 44.06 & 60.90  & 0.05 & 4.80 \\
    \textbf{DiffPen}         & 78.07 & 61.16 & 67.17 & 0.14 & 4.71 \\
    \textbf{Emuru}           & ~\textbf{9.85} & ~\textbf{4.33} & ~\textbf{5.60} & 0.11 & \textbf{1.28} \\
    \bottomrule
    \end{tabular}
    }\caption{Comparison on the line-level Karaoke dataset, separating calligraphy and typewritten styles, between Emuru and State-of-the-Art approaches trained on IAM. The KID is multiplied by $10^3$ and the best performance is in bold for readability.}
    \label{tab:typewritten}
    \vspace{-0.3cm}
\end{table}

\tit{Zero-shot inference on CVL and RIMES}
The dataset-specific tailoring of existing methods severely limits their generalization performance, as demonstrated in~\Cref{tab:cvl_rimes}. In this zero-shot evaluation on CVL and RIMES lines, all competing models (which have been trained on IAM) show a marked performance drop. In contrast, Emuru achieves top performance across nearly all metrics. Interestingly, despite not being able to capture the input style, some approaches maintain readability in the generated text, as reflected by the relatively stable $\Delta$CER score. These results, in line with those achieved on IAM, demonstrate the generalization capabilities of Emuru, also observed from~\Cref{fig:qualitative_comparison}.

\tit{Zero-shot inference on the Karaoke}
Finally, we evaluate the generalization capability of all models on calligraphy and typewritten text images by considering our devised Karaoke dataset. Different from IAM, CVL, and RIMES, Karaoke has a diverse set of 100 fonts and no background artifacts.  
In~\Cref{tab:typewritten}, we evaluate Emuru against the considered competitors on Karaoke, distinguishing between images of text in calligraphy fonts and typewritten fonts. 
In both cases, the results are in line with the zero-shot inference comparisons in~\Cref{tab:cvl_rimes}, with Emuru achieving the best values across all scores except $\Delta$CER. Arguably, this is because Emuru, like the diffusion-based models, does not have any mechanism to ensure the correctness of the generated text. In the Supplementary, we report further experiments, including fine-tuning Emuru with an HTR-based loss. Across all quantitative experiments in~\Cref{tab:iam,tab:cvl_rimes,tab:typewritten} and the qualitative comparison in~\Cref{fig:qualitative_comparison}, Emuru stands out as the only model that exhibits a consistent value range for the scores and comparable zero-shot performance on all the considered datasets. On the other hand, other methods show significant variability, sometimes completely failing to generate images in the style of the target dataset. This highlights Emuru's robustness and ability to generalize to diverse scenarios. 

\begin{figure}
    \centering
    \includegraphics[width=\columnwidth]{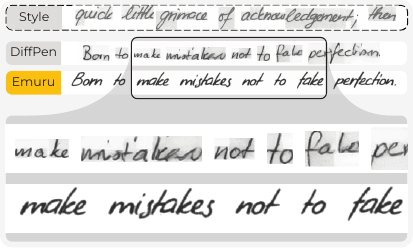}
    \caption{Results of image generation from a style image with background artifacts obtained with Emuru and DiffPen. Thanks to the VAE training, Emuru images do not reproduce these artifacts.}
    \label{fig:background}\vspace{-0.2cm}
\end{figure}

\tit{Editing Application}
As mentioned in previous sections, we train Emuru to generalize on unseen styles and to be background-agnostic. Combined, these two characteristics enable Emuru to generate images that are both faithful to the input style and free from background artifacts, which makes them usable in real-world editing scenarios. To illustrate this, in~\Cref{fig:overview}, we extract a style sample from a handwritten letter image and task Emuru to render the sentence ``\textit{Hello World! I am Emuru: Born to make mistakes, not to fake perfection.}''. Then, we overlay the resulting image on the same letter. Note that the same results cannot be replicated with other models due to the background artifacts they maintain, as shown in~\Cref{fig:background}. DiffPen effectively copies the style of the IAM sample, but it also reproduces the gray background typical of IAM samples. This issue of learning to imitate the background in addition to the text strokes is common in many HTG approaches because they are trained to mimic the training text images as a whole, including their background. In contrast, by leveraging the VAE's background removal capabilities, Emuru generates images that reproduce only the text style, and thus, that can be easily blended over any custom background image.

\section{Conclusions}
\label{sec:conclusions}
In this paper, we have presented Emuru, a novel autoregressive model for Offline styled text image generation. 
Our approach relies on a powerful text images representation model (a VAE) combined with an autoregressive Transformer, and both are trained on a diverse, synthetic dataset of English text rendered in over 100,000 typewritten and calligraphy fonts. 
Emuru effectively addresses the limitations of previous approaches, such as: generalization to novel styles, limited or fixed output length, and entanglement between text style and background style.
Extensive evaluation considering multiple datasets, both handwritten and typewritten, demonstrates the effectiveness of Emuru in generating both short and long text sequences with high fidelity to unseen styles and its superiority with respect to the State-of-the-Art in terms of generalization capabilities.

\clearpage
{
    \small
    \bibliographystyle{ieeenat_fullname}
    \bibliography{main}
}

\clearpage

\setcounter{page}{1}

\maketitlesupplementary

\appendix

{
  \hypersetup{linkcolor=black}
  \tableofcontents
}
\vspace{.5cm}

\resumetoc
\section{Further Details about the Training Dataset}
\label{sec:f2v2_details}

\begin{figure*}[t]
    \centering
    \includegraphics[width=\linewidth]{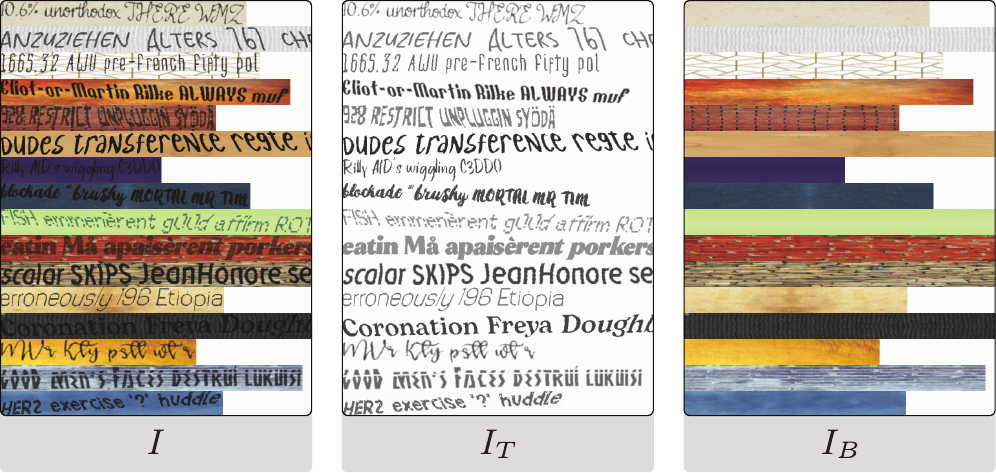}
    \caption{Exemplar sample images $I$ from the synthetic dataset used for training Emuru (left). These samples consist of a rendered greyscale text image $I_T$ (center) superimposed to RGB paper-like background images $I_B$ (right).}
    \label{fig:font2v2}\vspace{-.05cm}
\end{figure*}

The proposed synthetic dataset\footnote{\scriptsize{\url{https://huggingface.co/datasets/blowing-up-groundhogs/font-square-v2}}} comprises RGB images that feature text lines superimposed on diverse background patterns (\Cref{fig:font2v2}). The process to create a text image is divided into two macro-steps: Text Content Sampling and Text Line Rendering. 

\begin{figure}
    \centering
    \includegraphics[width=\columnwidth]{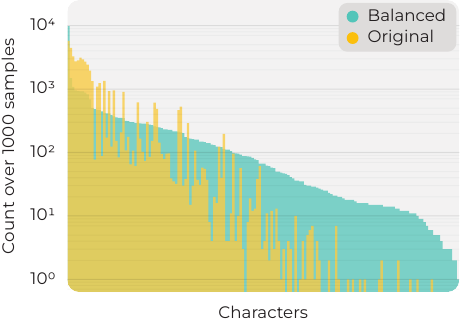}
    \caption{Characters frequency in text from the original corpora used to generate the synthetic training dataset, and from the balanced version of the corpora.}\vspace{-.05cm}
    \label{fig:char_dist}
\end{figure}

\tit{Text Content Sampling}
First, we formed the text lines to render in the samples of the dataset. To this end, we consider the following English corpora available on the NLTK\footnote{\scriptsize{\url{https://www.nltk.org/}}} library: \texttt{abc}, \texttt{brown}, \texttt{genesis}, \texttt{inaugural}, \texttt{state\_union}, and \texttt{webtext}. 
To ensure a balanced character distribution and to address the learning of rare characters, we employ a rarity-based weighting strategy. Specifically, each word in the corpora is assigned a weight, allowing for weighted sampling from the resulting distribution. The weight of each word is computed based on the frequency of unigrams and bigrams contained within it. This approach prioritizes words that include infrequent character patterns, enabling the HTG model to learn from the long-tail distribution of characters more effectively. 
The Python implementation of the algorithm for computing the weight of each word is reported in~\Cref{lst:balancing_procedure}. The function \texttt{word\_weight(word)} calculates a score for each word based on the unigram and bigram frequencies, stored in \texttt{u\_counts} and \texttt{b\_counts}, respectively. These counts represent the frequencies of unigrams and bigrams across the entire corpus. The resulting word weight is the average of the unigram and bigram-based scores. 
In \Cref{fig:char_dist}, we show the character's distribution with the custom weights per word and the naive sampling (original). Each bar of the distribution represents a character in our charset (which contains 157 characters) in this order:
[\texttt{~}, \texttt{e}, \texttt{a}, \texttt{n}, \texttt{s}, \texttt{i}, \texttt{r}, \texttt{t}, \texttt{o}, \texttt{l}, \texttt{A}, \texttt{.}, \texttt{m}, \texttt{c}, \texttt{d}, \texttt{u}, \texttt{0}, \texttt{S}, \texttt{h}, \texttt{E}, \texttt{p}, \texttt{M}, \texttt{g}, \texttt{C}, \texttt{R}, \texttt{1}, \texttt{I}, \texttt{-}, \texttt{D}, \texttt{b}, \texttt{T}, \texttt{2}, \texttt{N}, \texttt{3}, \texttt{O}, \texttt{y}, \texttt{P}, \texttt{B}, \texttt{L}, \texttt{'}, \texttt{F}, \texttt{k}, \texttt{f}, \texttt{H}, \texttt{4}, \texttt{5}, \texttt{G}, \texttt{U}, \texttt{v}, \texttt{7}, \texttt{8}, \texttt{"}, \texttt{6}, \texttt{9}, \texttt{V}, \texttt{K}, \texttt{x}, \texttt{w}, \texttt{Y}, \texttt{)}, \texttt{ä}, \texttt{W}, \texttt{,}, \texttt{*}, \texttt{z}, \texttt{:}, \texttt{J}, \texttt{(}, \texttt{X}, \texttt{!}, \texttt{j}, \texttt{/}, \texttt{\_}, \texttt{é}, \texttt{>}, \texttt{Z}, \texttt{?}, \texttt{Q}, \texttt{;}, \texttt{\$}, \texttt{å}, \texttt{]}, \texttt{ö}, \texttt{q}, \texttt{\%}, \texttt{<}, \texttt{[}, \texttt{á}, \texttt{ã}, \texttt{\#}, \texttt{¡}, \texttt{\&}, \texttt{\}}, \texttt{+}, \texttt{ü}, \texttt{â}, \texttt{ó}, \texttt{Ą}, \texttt{ê}, \texttt{=}, \texttt{\{}, \texttt{ô}, \texttt{\textbackslash{}}, \texttt{í}, \texttt{è}, \texttt{»}, \texttt{╜}, \texttt{É}, \texttt{ú}, \texttt{@}, \texttt{À}, \texttt{à}, \texttt{\`}, \texttt{ç}, \texttt{ù}, \texttt{\~{}}, \texttt{¨}, \texttt{´}, \texttt{§}, \texttt{˝}, \texttt{µ}, \texttt{“}, \texttt{Ă}, \texttt{č}, \texttt{°}, \texttt{Â}, \texttt{®}, \texttt{ŕ}, \texttt{—}, \texttt{¢}, \texttt{Ä}, \texttt{û}, \texttt{Ó}, \texttt{”}, \texttt{ƒ}, \texttt{Ť}, \texttt{\^{}}, \texttt{ï}, \texttt{˘}, \texttt{|}, \texttt{î}, \texttt{ë}, \texttt{¦}, \texttt{Ś}, \texttt{õ}, \texttt{Ö}, \texttt{Á}, \texttt{º}, \texttt{Å}, \texttt{ť}, \texttt{Ł}, \texttt{Ü}, \texttt{ß}, \texttt{ż}, \texttt{ñ}, \texttt{ĺ}, \texttt{ď}].
\begin{lstlisting}[float,caption={Python code for computing the words weight, needed for the character frequency balancing procedure adopted to obtain the text in our synthetic training dataset.},label={lst:balancing_procedure},captionpos=b,floatplacement=t]
def word_weight(word):
    # Compute the score
    # based on the unigrams
    u_score = 0
    for c in word:
        u_score += u_counts[c]
    u_score /= len(word)
    
    # Compute the score
    # based on the bigrams
    bigrams = pairwise(f' {word} ')
    bigrams = list(bigrams)
    b_score = 0
    for b in bigrams:
        b_score += b_counts[''.join(b)]
    b_score /= len(bigrams)

    # Average the two scores
    return (u_score + b_score) / 2 
\end{lstlisting}
Note that, our current Emuru is trained on a dataset derived from fonts covering primarily Latin characters, there is nothing in the approach itself that prohibits extension to non-Latin scripts. 
Specifically, to extend Emuru to non-Latin scripts, its VAE should be re-trained on a similarly-built synthetic dataset of non-Latin strings, rendered with fonts that support the desired characters. Given that there are a number of such fonts available online, we argue that there is no obvious method or model limitation as to why the results of Emuru would not translate also to non-Latin scripts.

\tit{Text Line Rendering}
Once we sample a sentence, we select a font from a list of 100,000 fonts and render the text image on a white background, obtaining $\hat{I}_T$. Then, we sample a background $\hat{I}_B$ from a set of images depicting reasonable supports for writings (\eg paper-like textures, wood, walls), a transparency value $\alpha \in [0.5, 1]$ and process $\hat{I}_T$ and $\hat{I}_B$ in parallel. To obtain $I_T$ (\Cref{fig:font2v2}, center), we use the following random transforms: Rotation, Warping, Gaussian Blur, Dilation, and Color Jitter (with probabilities respectively set to 0.5, 1.0, 0.5, 0.1, and 0.5) and apply the transparency $\alpha$. To obtain $I_B$ (\Cref{fig:font2v2}, right), we use the following random transforms: Dilation, Color Jitter, and Random Inversion of the values (with probabilities respectively set to 0.1, 0.5, 0.2). The image $I$ (\Cref{fig:font2v2}, left) is the result of superimposing $I_T$ on $I_B$. 
\section{Details about the Considered Datasets}
\label{sec:datasets_details}
In this section, we give further details about the datasets we use for evaluation.
\tit{IAM}
The IAM Handwriting Database~\cite{marti2002iam} is a collection of greyscale document scans written in English by multiple writers. The dataset features free-layout modern English text lines from the Lancaster-Oslo/Bergen (LOB) corpus. The number of pages per writer varies significantly, ranging from 1 to 10. The dataset comes with both line-level and word-level segmentation information. The HTG community commonly used this dataset for training and evaluation of the proposed approaches and has adopted a split defined in~\cite{kang2020ganwriting}, in which the samples from 339 writers are used for training, and those from 161 other writers are used for test. In this work, we adopt the same split. 
However, while most of the competitors are trained on IAM (and, therefore, see the test set as in-distribution samples), we never use any training data from it to train our model (therefore, test samples are out-of-distribution for us).
\tit{CVL} 
The CVL Database~\cite{kleber2013cvl} is a collection of RGB scans of English and German manuscripts written with ink on white paper by 310 writers. In this work, we consider the line-level annotation of the dataset and perform the evaluation on its test set, which contains lines from 283 different authors. We include this dataset since it is mostly in the same language as the commonly-adopted IAM, but the handwriting styles are out-of-distribution \wrt~IAM.
\tit{RIMES} 
The RIMES Database~\cite{augustin2006rimes} is a collection of binary images of customer service-themed letters in French, written by multiple authors. For our experiments, we consider the lines in the official test split of the dataset, which have been written by 100 authors. We include this dataset as a more challenging collection of out-of-distribution generation cases. In fact, both the language and the styles differ from those in IAM.
\tit{Karaoke}\footnote{\scriptsize{\url{https://huggingface.co/datasets/blowing-up-groundhogs/karaoke}, \url{https://open.spotify.com/playlist/5GTPvzVNedUIkYkV7xOtgI}}}
 To assess the capabilities of Emuru and the existing handwriting style imitation approaches, we devise a dataset of text lines images obtained by rendering song lyrics with 100 publicly available fonts\footnote{\scriptsize{\url{https://fonts.google.com/}}}, 50 calligraphy, and 50 typewritten, on a white background. This dataset, especially its typewritten split, serves as a challenging test to measure the performance of HTG models when dealing with significantly out-of-distribution, font-like styles.

\section{Details about the Considered Scores}
\label{sec:scores_details}
Measuring the performance of styled HTG models is a challenging task per se. Early works on HTG simply employ the Fréchet Inception Distance (FID) to measure the realism of the generated images, and sometimes the CER to measure their readability. 
Recent works~\cite{nikolaidou2024rethinking, vanherle2024vatr++, pippi2023hwd} have highlighted the limitations of such scores (especially the FID) and thus proposed to assess the performance of HTG models both by using multiple scores or introducing novel, task-specific scores. In line with these works, for our evaluation, we adopt the following scores (some of which have been reported in the main paper).
\tit{Fréchet Inception Distance (FID)} 
The FID~\cite{heusel2017gans} is widely employed in HTG literature and captures the realism/similarity between the distribution of the real images and the generated ones. In the context of HTG, it somehow captures the texture-wise style similarity. Note that in most HTG works, the reported FID is computed only on the initial square crop of the real and generated images. In this work, we compute it on non-overlapping crops obtained from the entire images.
\tit{Kernel Inception Distance (KID)}
Similar to the FID, also the KID~\cite{binkowski2018demystifying} captures the realism of the generated images by comparing their distribution to the real images distribution. Despite being less used in image generation evaluation, it is more numerically stable than the FID~\cite{pippi2023hwd}. 
\tit{Binarized FID (BFID)}
To reduce the impact of the texture (both of the strokes and the background) on the similarity between generated and real images, we introduce this variant of the FID score by computing it on binarized images. Intuitively this measure should capture mostly the perceptual handwriting style and disregard color or textures. 
\tit{Binarized KID (BKID)}
Similar to the BFID, this score is obtained by computing the KID on the binarized reference and generated images. 
\tit{Absolute Character Error Rate Difference ($\Delta$CER)}
We consider the State-of-the-Art TrOCR-Base~\cite{li2021trocr} model, which can handle both handwritten and typewritten images and compute its CER on the reference and generated images to capture their readability while taking into consideration the style. Intuitively, we want the generated images to be as readable as the reference ones (which suggest that they actually contain text) but not significantly more, which could be that the HTG model has collapsed to a very readable style that does not necessarily resemble the reference.
\tit{Handwriting Distance (HWD)} 
The HWD~\cite{pippi2023hwd} is a perceptual score that has been recently proposed specifically for HTG and captures the similarity in terms of calligraphic style between the real and generated images pairs. For this reason, we include it in our evaluation setup.
\tit{Absolute Intra Learned Perceptual Image Patch Similarity Difference ($\Delta$I-LPIPS)} 
To measure the style preservation over long generated images, we split each image into non-overlapping crops and compute the LPIPS~\cite{zhang2018unreasonable} (which is based on the similarity of feature maps) between all possible combinations of crops of the same image. We repeat this process for both the real and the generated image separately and then compute the absolute difference between these values. We want its value on the generated images to be similar to what is obtained on the real ones (to ensure mimicking the degree of variability of the reference style) and not significantly lower, which could indicate mode collapse to a repeated stroke.
\begin{figure}
    \centering
    \includegraphics[width=\columnwidth]{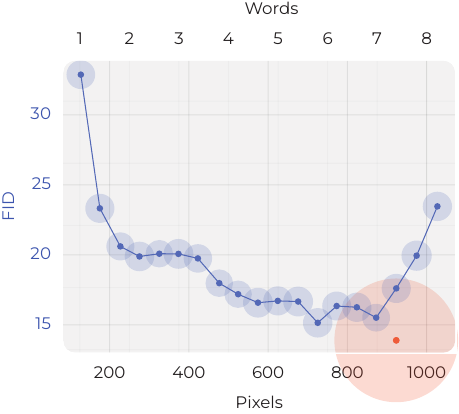}
    \caption{Emuru FID scores for IAM line generation using style inputs of varying length (in blue), compared to those generated with the original input style (in red). The length of the style input is expressed as the average pixel width of the style samples, with standard deviations represented by semi-transparent circles behind the data points. Additionally, the average number of words in each style sample is indicated, where an ``average word'' corresponds to approximately 125 pixels in width.}
    \label{fig:score_wrt_inlen}
\end{figure}
\section{Analysis on the Style Input Length}
\label{sec:input_study}
As mentioned in~\Cref{sec:related}, 
existing HTG methods have a variety of settings for the style reference image, ranging from single words~\cite{gan2021higan, gan2022higan+, krishnan2023textstylebrush, dai2024one} to entire paragraphs~\cite{mayr2024zero}. These approaches offer different advantages depending on the use case, with some models demonstrating some adaptability to a style from minimal input and others leveraging more context for improved style fidelity. We argue that an HTG model should be flexible to the number of input words available, handling both data-scarce scenarios with limited reference samples and situations with more style images available. To deepen the analysis on this aspect, we evaluate the performance of Emuru when generating IAM lines given increasingly longer reference style images, which we build starting from the word-level annotations of the IAM dataset. Note that the IAM dataset contains all the information needed to determine which sequence of words appears within each line and where to cut the line image to obtain a new line image with the desired width. Therefore, to generate the set of style samples within a specific width range, we exploit the IAM information of the bounding boxes around the words inside the lines. Specifically, we group images whose width is in a range of $\pm$25px around average width values that vary from a minimum of 125px to a maximum of 1025px with steps of 50px. 
In~\Cref{fig:score_wrt_inlen}, we report the results of this analysis. As expected, the performance improves with increasing input length, showing that our model is flexible to different scenarios and more applicable to real-world handwriting generation tasks. 

\section{Analysis on the Styled Output Length}
\label{sec:output_study}
\begin{figure}[]
    \centering
    \begin{tabular}{c}
    \\
         \includegraphics[width=\columnwidth]{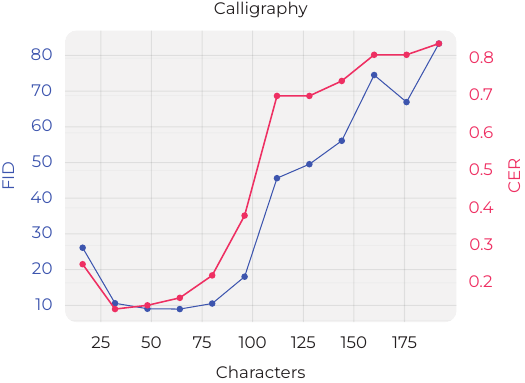}
         \\ \\
         \includegraphics[width=\columnwidth]{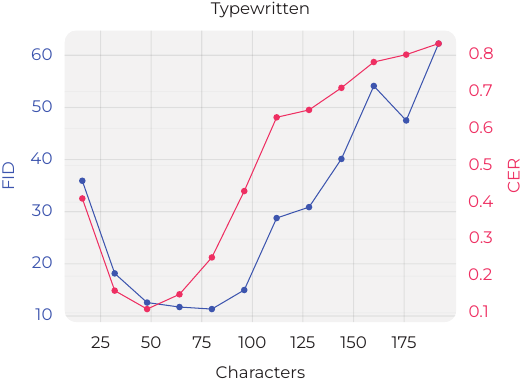}
    \end{tabular}
    \caption{Emuru FID and CER when generating increasingly longer lines in unseen styles from the Karaoke dataset, separated into calligraphy and typewritten styles. We express the generated lines length in terms of the number of characters contained.} \vspace{-.05cm}
    \label{fig:score_wrt_outlen}
\end{figure}
\begin{landscape}
    \begin{figure}
    \vspace{-.6cm}
        \centering
        \begin{tabular}{cc}
            \includegraphics[width=\textwidth]{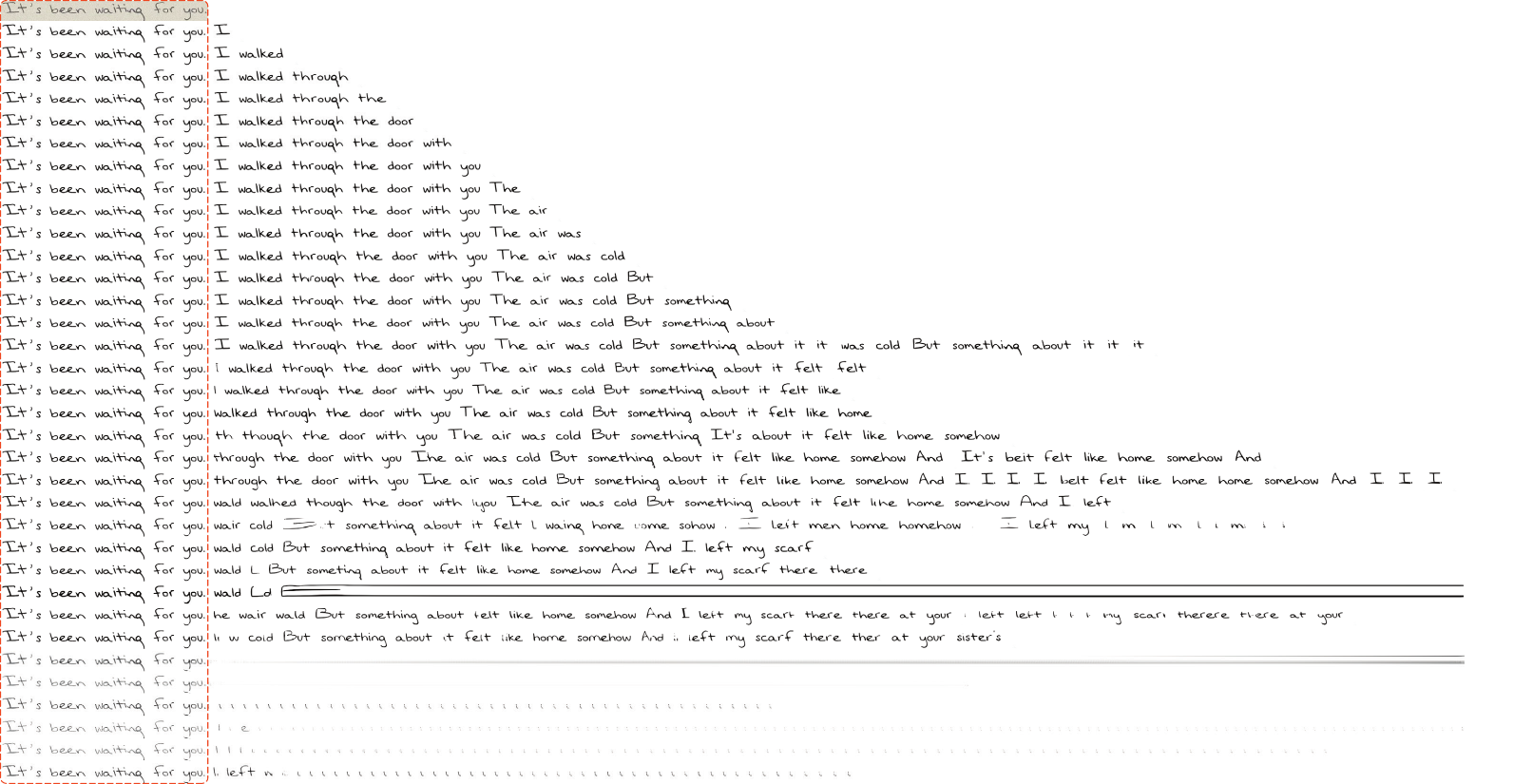} \\ \\
            \includegraphics[width=\textwidth]{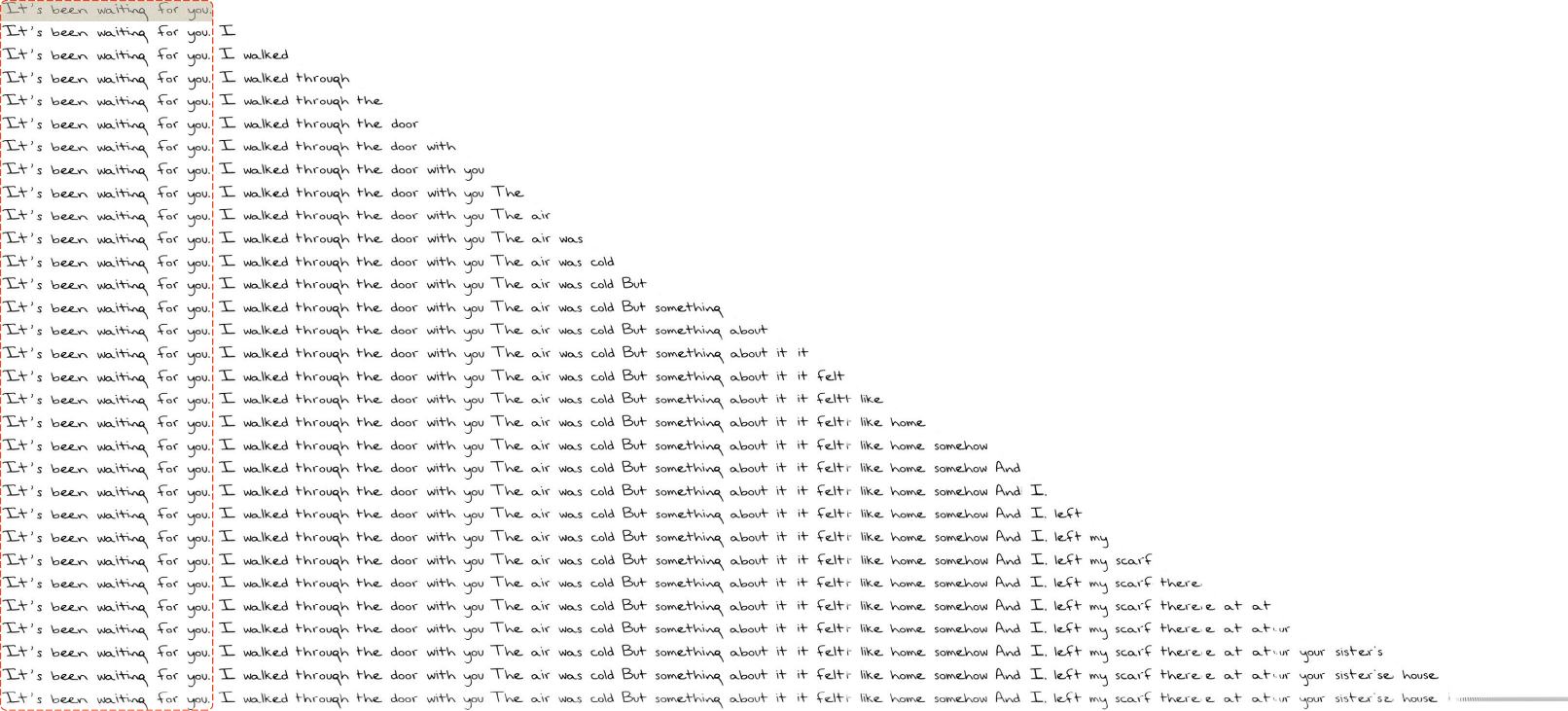} 
        \end{tabular}
        \caption{Increasingly long text image generation with Emuru. We use the top, colored image as style input and task the model to generate an incrementally longer sentence in each line. In the top figure, we task the model to generate the whole sentence in one go, while, in the bottom one, each line is used as the style sample of the next one and the model generates only the last word.}
        \label{fig:qualitaitve_outlen}
    \end{figure}
\end{landscape}

To assess the optimal output text length manageable by Emuru, we perform experiments by increasing the number of output characters to render in the styles from the Karaoke dataset. We report FID and CER scores in~\Cref{fig:score_wrt_outlen}, representing style fidelity and text readability. We observe that the model performs best when generating text lines between 25 and 75 characters, which aligns with the training sequence length. Outside this range, Emuru exhibits performance degradation in both style fidelity and text readability. Note that some workarounds for improving the quality of generated long sequences can be applied, as mentioned in the following. 
In~\Cref{fig:qualitaitve_outlen}, we report qualitative results of image generation with Emuru for increasingly longer text lines. In the top part of the figure, we feed Emuru with the reference style image and the entire text line to be generated all at once. As we can see, the model maintains style consistency across long outputs but struggles to correctly render all the words as their number increases. This discrepancy between the input and the output text is reflected in the $\Delta$CER score, as observed in the quantitative analysis in Section 4~\Cref{sec:experiments}. 
To improve this aspect, in the bottom part of~\Cref{fig:qualitaitve_outlen}, we also show a simple yet effective solution for generating longer lines with increased text fidelity. Specifically, we let Emuru iteratively generate one word at a time by using the previously generated line as a style reference. This incremental approach improves the alignment of the output image with the input text with a computational overhead of 29.5\% (2.24s), mainly due to the additional 10 padding embeddings that the model has to generate as a stop signal after each new word generation.
\section{Detailed Comparison Between VAEs}
\label{sec:vaes_details}
In this section we expand the analysis on the performance of Emuru VAE on multiple datasets reported in~\Cref{sec:experiments}. 
As the performance of the autoregressive Transformer in Emuru is tied to the quality of the latent vectors it has been trained on, this detailed analysis helps determine a soft bound on the reconstruction quality.
In~\Cref{tab:best_vae_supp}, we report the results of the reconstruction performance comparison on different datasets, which we compute on images reconstructed with different VAEs: Emuru VAE, SD1.5 VAE (used by DiffPen and OneDM), and SD3 VAE (the current State-of-the-Art in image reconstruction). As we can see, Emuru VAE consistently gives the best or second-best BFID and BKID due to its ability to closely capture and reproduce the handwritten style. Looking at FID and KID, in some datasets the performance is a bit lower \wrt the alternatives considered, likely due to the fact that Emuru VAE is trained to reconstruct text on a clean background (and not the original input image). Moreover, the $\Delta$I-LPIPS is consistently better, which indicates that the images reproduced by our VAE have a consistent style in all their parts. In general, we can observe either better or comparable performance to the other considered VAEs. The tailored training of our VAE allows us to obtain these results with a much more lightweight model and a more compressed latent space ($\sim$16\% of the parameters of the other VAEs and 1 latent channel instead of 4 and 16, respectively for SD1.5 VAE and SD3 VAE). 

\begin{table}[]
    \centering
    \setlength{\tabcolsep}{.18em}
    \resizebox{\columnwidth}{!}{%
    \begin{tabular}{l ccccccc}
    \toprule
    & \multicolumn{7}{c}{\textbf{IAM words}}\\ 
    \cmidrule{2-8}
    & \textbf{FID$\downarrow$}
    & \textbf{BFID$\downarrow$}
    & \textbf{KID$\downarrow$}
    & \textbf{BKID$\downarrow$}
    & \textbf{$\Delta$CER$\downarrow$}
    & \textbf{HWD$\downarrow$} 
    & \textbf{$\Delta$I-LPIPS$\downarrow$}\\
    \midrule
    \textbf{SD1.5 VAE}  & 21.15 & 2.42 & 20.72 & 1.65 & \textbf{0.01} & 0.92 & 2.34 \\
    \textbf{SD3 VAE}    & \textbf{15.28} & \textbf{2.09} & \textbf{13.62} & 1.35 & \textbf{0.01} & 0.84 & 2.72 \\
    \textbf{Emuru VAE}  & 27.74 & 2.60 & 23.41 & \textbf{1.32} & 0.08 & \textbf{0.64} & \textbf{0.02} \\   
    \midrule
    & \multicolumn{7}{c}{\textbf{IAM lines}}\\ 
    \cmidrule{2-8}
    & \textbf{FID$\downarrow$}
    & \textbf{BFID$\downarrow$}
    & \textbf{KID$\downarrow$}
    & \textbf{BKID$\downarrow$}
    & \textbf{$\Delta$CER$\downarrow$}
    & \textbf{HWD$\downarrow$} 
    & \textbf{$\Delta$I-LPIPS$\downarrow$}\\
    \midrule
    \textbf{SD1.5 VAE}  & 15.87 & 1.64 & 16.47 & 1.27 & 0.01 & 0.68 & 74.04 \\
    \textbf{SD3 VAE}    & ~\textbf{9.80} & 1.77 & ~\textbf{9.66} & 1.53 & \textbf{0.00} & \textbf{0.59} & 82.95 \\
    \textbf{Emuru VAE}  & 16.74 & \textbf{1.11} & 13.76 & \textbf{0.53} & 0.01 & 0.76 & ~\textbf{6.24} \\
    \midrule
    & \multicolumn{7}{c}{\textbf{CVL}}\\ 
    \cmidrule{2-8}
    & \textbf{FID$\downarrow$}
    & \textbf{BFID$\downarrow$}
    & \textbf{KID$\downarrow$}
    & \textbf{BKID$\downarrow$}
    & \textbf{$\Delta$CER$\downarrow$}
    & \textbf{HWD$\downarrow$} 
    & \textbf{$\Delta$I-LPIPS$\downarrow$}\\
    \midrule
    \textbf{SD1.5 VAE}  & 83.33 & 34.82 & 101.89 & 34.74 & 0.01 & 0.63 & 45.74 \\
    \textbf{SD3 VAE}    & 70.84 & ~8.77 & ~84.68 & ~9.30 & \textbf{0.00} & \textbf{0.57} & 52.81 \\
    \textbf{Emuru VAE}  & \textbf{11.25} & ~\textbf{1.10} & ~~\textbf{9.77} & ~\textbf{0.53} & 0.05 & 0.79 & ~\textbf{5.28} \\
    \midrule
    & \multicolumn{7}{c}{\textbf{RIMES}}\\ 
    \cmidrule{2-8}
    & \textbf{FID$\downarrow$}
    & \textbf{BFID$\downarrow$}
    & \textbf{KID$\downarrow$}
    & \textbf{BKID$\downarrow$}
    & \textbf{$\Delta$CER$\downarrow$}
    & \textbf{HWD$\downarrow$} 
    & \textbf{$\Delta$I-LPIPS$\downarrow$}\\
    \midrule
    \textbf{SD1.5 VAE}  & 20.35 & 2.77 & 19.57 & 2.14 & 0.02 & \textbf{0.79} & ~\textbf{0.50} \\
    \textbf{SD3 VAE}    & \textbf{19.07} & 5.10 & \textbf{17.65} & 4.95 & 0.03 & 0.90 & ~3.52 \\
    \textbf{Emuru VAE}  & 47.79 & \textbf{2.62} & 44.27 & \textbf{1.45} & \textbf{0.01} & 1.24 & 39.72 \\  
    \midrule
    & \multicolumn{7}{c}{\textbf{Karaoke Typewritten}}\\ 
    \cmidrule{2-8}
    & \textbf{FID$\downarrow$}
    & \textbf{BFID$\downarrow$}
    & \textbf{KID$\downarrow$}
    & \textbf{BKID$\downarrow$}
    & \textbf{$\Delta$CER$\downarrow$}
    & \textbf{HWD$\downarrow$} 
    & \textbf{$\Delta$I-LPIPS$\downarrow$}\\
    \midrule
    \textbf{SD1.5 VAE}  & 12.28 & 0.87 & 9.54 & 0.31 & \textbf{0.01} & \textbf{0.55} & 20.60 \\
    \textbf{SD3 VAE}    & ~\textbf{5.29} & 1.13 & \textbf{2.89} & 0.50 & \textbf{0.01} & 0.65 & 32.76 \\
    \textbf{Emuru VAE}  & ~6.55 & \textbf{0.75} & 3.92 & \textbf{0.09} & \textbf{0.01} & 0.56 & ~\textbf{7.43} \\ 
    \midrule
    & \multicolumn{7}{c}{\textbf{Karaoke Calligraphy}}\\ 
    \cmidrule{2-8}
    & \textbf{FID$\downarrow$}
    & \textbf{BFID$\downarrow$}
    & \textbf{KID$\downarrow$}
    & \textbf{BKID$\downarrow$}
    & \textbf{$\Delta$CER$\downarrow$}
    & \textbf{HWD$\downarrow$} 
    & \textbf{$\Delta$I-LPIPS$\downarrow$}\\
    \midrule
    \textbf{SD1.5 VAE}  & 23.38 & 1.63 & 24.65 & 1.16 & \textbf{0.00} & 1.04 & ~\textbf{9.37} \\
    \textbf{SD3 VAE}    & 11.09 & 2.80 & ~9.58 & 2.37 & \textbf{0.00} & \textbf{0.87} & 13.84 \\
    \textbf{Emuru VAE}  & ~\textbf{5.28} & \textbf{1.55} & ~\textbf{2.98} & \textbf{0.51} & \textbf{0.00} & 1.13 & 10.63 \\ 
    \bottomrule
    \end{tabular}
     }
    \caption{Comparison between the VAE featured in Emuru and other VAEs used in popular Diffusion Models when reproducing the images in the test set of different HTG datasets. Note that none of these VAEs has been trained on the considered datasets. KID, BKID, and $\Delta$I-LPIPS are multiplied by $10^3$ and the best performance is in bold for readability.}
    \label{tab:best_vae_supp}
\end{table}

\section{Multi-stage Training Ablation}
\label{sec:training_ablation}
\begin{table}[t]
    \centering
    \setlength{\tabcolsep}{.18em}
    \resizebox{\columnwidth}{!}{%
    \begin{tabular}{l ccccccc}
    \toprule
    & \multicolumn{7}{c}{\textbf{IAM words}}\\ 
    \cmidrule{2-8}
    & \textbf{FID$\downarrow$}
    & \textbf{BFID$\downarrow$}
    & \textbf{KID$\downarrow$}
    & \textbf{BKID$\downarrow$}
    & \textbf{$\Delta$CER$\downarrow$}
    & \textbf{HWD$\downarrow$} 
    & \textbf{$\Delta$I-LPIPS$\downarrow$}\\
    \midrule
    \textbf{Pretraining}        & 85.15 & 53.28 & 83.53 & 52.54 & 0.56 & 3.51 & 0.39 \\
    \textbf{+ Var. len. ft}     & \textbf{63.61} & \textbf{37.73} & \textbf{62.34} & \textbf{37.22} &\textbf{ 0.19} & \textbf{3.03} & \textbf{0.16} \\
    \midrule
    & \multicolumn{7}{c}{\textbf{IAM lines}}\\ 
    \cmidrule{2-8}
    & \textbf{FID$\downarrow$}
    & \textbf{BFID$\downarrow$}
    & \textbf{KID$\downarrow$}
    & \textbf{BKID$\downarrow$}
    & \textbf{$\Delta$CER$\downarrow$}
    & \textbf{HWD$\downarrow$} 
    & \textbf{$\Delta$I-LPIPS$\downarrow$}\\
    \midrule
    \textbf{Pretraining}        & 44.90 & 42.36 & 28.19 & 29.75 & 0.83 & 2.70 & \textbf{11.39} \\
    \textbf{+ Var. len. ft}     & \textbf{13.89} & \textbf{~6.19} & \textbf{11.30} & \textbf{~5.36} & \textbf{0.14} & \textbf{1.87} & 38.27 \\
    \midrule
    & \multicolumn{7}{c}{\textbf{CVL}}\\ 
    \cmidrule{2-8}
    & \textbf{FID$\downarrow$}
    & \textbf{BFID$\downarrow$}
    & \textbf{KID$\downarrow$}
    & \textbf{BKID$\downarrow$}
    & \textbf{$\Delta$CER$\downarrow$}
    & \textbf{HWD$\downarrow$} 
    & \textbf{$\Delta$I-LPIPS$\downarrow$}\\
    \midrule
    \textbf{Pretraining}        & 77.23 & 70.52 & 69.28 & 54.47 & 0.87 & 2.52 & 6.69 \\
    \textbf{+ Var. len. ft}     & \textbf{14.39} & \textbf{10.77} & \textbf{12.34} & \textbf{10.54} & \textbf{0.13} & \textbf{1.82} & \textbf{0.75} \\
    \midrule
    & \multicolumn{7}{c}{\textbf{RIMES}}\\ 
    \cmidrule{2-8}
    & \textbf{FID$\downarrow$}
    & \textbf{BFID$\downarrow$}
    & \textbf{KID$\downarrow$}
    & \textbf{BKID$\downarrow$}
    & \textbf{$\Delta$CER$\downarrow$}
    & \textbf{HWD$\downarrow$} 
    & \textbf{$\Delta$I-LPIPS$\downarrow$}\\
    \midrule
    \textbf{Pretraining}        & 100.34 & 75.90 & 76.32 & 56.07 & 0.69 & 2.96 & 87.25 \\
    \textbf{+ Var. len. ft}     & ~\textbf{26.93} & \textbf{13.36} & \textbf{21.19} & \textbf{~9.40} & \textbf{0.25} & \textbf{2.18} & \textbf{47.96} \\
    \midrule
    & \multicolumn{7}{c}{\textbf{Karaoke Typewritten}}\\ 
    \cmidrule{2-8}
    & \textbf{FID$\downarrow$}
    & \textbf{BFID$\downarrow$}
    & \textbf{KID$\downarrow$}
    & \textbf{BKID$\downarrow$}
    & \textbf{$\Delta$CER$\downarrow$}
    & \textbf{HWD$\downarrow$} 
    & \textbf{$\Delta$I-LPIPS$\downarrow$}\\
    \midrule
    \textbf{Pretraining}        & 47.06 & 39.28 & 22.76 & 14.69 & 0.67 & 2.03 & 45.52 \\
    \textbf{+ Var. len. ft}     & \textbf{~9.85} & \textbf{~4.33} & \textbf{~5.60} & \textbf{~1.24} & \textbf{0.11} & 1.28 & ~\textbf{5.07} \\
    \midrule
    & \multicolumn{7}{c}{\textbf{Karaoke Calligraphy}}\\ 
    \cmidrule{2-8}
    & \textbf{FID$\downarrow$}
    & \textbf{BFID$\downarrow$}
    & \textbf{KID$\downarrow$}
    & \textbf{BKID$\downarrow$}
    & \textbf{$\Delta$CER$\downarrow$}
    & \textbf{HWD$\downarrow$} 
    & \textbf{$\Delta$I-LPIPS$\downarrow$}\\
    \midrule
    \textbf{Pretraining}        & 37.26 & 29.65 & 18.54 & 16.81 & 0.54 & 2.99 & ~4.90 \\
    \textbf{+ Var. len. ft}     & \textbf{13.87} & \textbf{~7.99} & \textbf{~9.24} & \textbf{~5.37} & \textbf{0.13} & \textbf{2.24} & ~\textbf{0.73} \\
    \bottomrule
    \end{tabular}
     }
    \caption{Ablation analysis on the multi-stage training strategy for the autoregressive Transformer (`ft' stands for `fine-tuning'). The scores are computed on all the considered datasets. KID, BKID, and $\Delta$I-LPIPS are multiplied by $10^3$ and the best performance is in bold for readability.}
    \label{tab:training}
\end{table}
In ~\Cref{tab:training}, we report a detailed analysis of the performance obtained with Emuru across different training stages. 
As explained in~\Cref{sec:method}, 
we train Emuru with the MSE loss in two stages: pretraining with text lines containing 4 to 7 words (\textbf{Pretraining} in~\Cref{tab:training}) and fine-tuning on text lines containing 1 to 32 words (\textbf{+ Var. len. ft} in ~\Cref{tab:training}). Predictably, there is a significant performance improvement across all metrics and datasets, which contain variable-length text lines. These results indicate the benefit of the two-stage training approach. 
During the first stage, the autoregressive Transformer learns to correlate the text input to the VAE embeddings extracted from the reference style image in a simpler setup. Then, in the second stage, the model learns to deal with variable-length inputs and can also focus on understanding when to stop generating. In fact, although the pertaining stage is longer, the samples used in this stage have similar text length and reference image width, and thus, batches of such samples are more informative since they contain less padding. On the other hand, the images used in the variable length fine-tuning stage need to be padded more to be batched, which makes this stage less efficient. Nonetheless, thanks to pertaining, few iterations are sufficient to achieve good performance.
\section{Additional Quantitative Results}
\label{sec:quantitatives_supplementary}
In~\Cref{tab:iam_supp,tab:cvl_rimes_supp,tab:karaoke_supp}, we report the performance of Emuru compared to State-of-the-Art HTG solutions on the considered datasets, expressed in terms of all the scores in the extended set of considered scores, as described in~\Cref{sec:scores_details}. 
\begin{table}[]
    \centering
    \setlength{\tabcolsep}{.18em}
    \resizebox{\columnwidth}{!}{%
    \begin{tabular}{l ccccccc}
    \toprule
    & \multicolumn{7}{c}{\textbf{IAM words}}\\ 
    \cmidrule{2-8}
    & \textbf{FID$\downarrow$}
    & \textbf{BFID$\downarrow$}
    & \textbf{KID$\downarrow$}
    & \textbf{BKID$\downarrow$}
    & \textbf{$\Delta$CER$\downarrow$}
    & \textbf{HWD$\downarrow$} 
    & \textbf{$\Delta$I-LPIPS$\downarrow$}\\
    \midrule
    \textbf{TS-GAN}          & 129.57 & 86.45 & 141.08 & 92.48 & 0.28 & 4.22 & 1.15 \\
    \textbf{HiGAN+}          & ~50.19 & 21.92 & ~43.39 & 14.21 & 0.20 & 3.12 & 0.17 \\
    \textbf{HWT}             & ~27.83 & 15.09 & ~19.64 & 11.95 & 0.15 & 2.01 & 0.06 \\
    \textbf{VATr}            & ~30.26 & 15.81 & ~22.31 & 13.37 & \textbf{0.00} & 2.19 & 0.56 \\
    \textbf{VATr++}          & ~31.91 & 17.15 & ~23.05 & 15.20 & 0.07 & 2.54 & 0.55 \\
    \textbf{One-DM}          & ~27.54 & 10.73 & ~21.39 & ~6.72 & 0.10 & 2.28 & \textbf{0.00} \\
    \textbf{DiffPen}         & ~\textbf{15.54} & ~\textbf{6.06} & ~\textbf{11.55} & ~\textbf{3.93} & 0.06 & \textbf{1.78} & 0.53 \\
    \textbf{Emuru}           & ~63.61 & 37.73 & ~62.34 & 37.22 & 0.19 & 3.03 & 0.16 \\  
    \midrule
    & \multicolumn{7}{c}{\textbf{IAM lines}}\\ 
    \cmidrule{2-8}
    & \textbf{FID$\downarrow$}
    & \textbf{BFID$\downarrow$}
    & \textbf{KID$\downarrow$}
    & \textbf{BKID$\downarrow$}
    & \textbf{$\Delta$CER$\downarrow$}
    & \textbf{HWD$\downarrow$} 
    & \textbf{$\Delta$I-LPIPS$\downarrow$}\\
    \midrule
    \textbf{TS-GAN}          & 44.17 & 19.45 & 45.42 & 18.17 & 0.02 & 3.21 & 25.66 \\
    \textbf{HiGAN+}          & 74.41 & 34.18 & 77.27 & 31.24 & \textbf{0.00} & 3.25 & 63.43 \\
    \textbf{HWT}             & 44.72 & 30.26 & 43.49 & 31.14 & 0.33 & 2.97 & 34.31 \\
    \textbf{VATr}            & 35.32 & 27.97 & 33.61 & 27.81 & 0.02 & 2.37 & 28.17 \\
    \textbf{VATr++}          & 34.00 & 21.67 & 29.68 & 19.04 & 0.03 & 2.38 & 35.86 \\
    \textbf{One-DM}          & 43.89 & 21.54 & 44.48 & 20.94 & 0.13 & 2.83 & 78.42 \\
    \textbf{DiffPen}         & \textbf{12.89} & ~6.87 & ~9.73 & ~\textbf{4.98} & 0.03 & 2.13 & ~\textbf{3.27} \\
    \textbf{Emuru}           & 13.89 & ~\textbf{6.19} & ~\textbf{1.30} & ~5.36 & 0.14 & \textbf{1.87} & 38.27 \\
    \bottomrule
    \end{tabular}
     }
    \caption{Comparison on the word-level and line-level IAM datasets between Emuru and State-of-the-Art approaches trained on IAM. KID, BKID, and $\Delta$I-LPIPS are multiplied by $10^3$ and the best performance is in bold for readability.}
    \label{tab:iam_supp}
\end{table}
\begin{table}[]
    \centering
    \setlength{\tabcolsep}{.18em}
    \resizebox{\columnwidth}{!}{%
    \begin{tabular}{l ccccccc}
    \toprule
    & \multicolumn{7}{c}{\textbf{CVL}}\\ 
    \cmidrule{2-8}
    & \textbf{FID$\downarrow$}
    & \textbf{BFID$\downarrow$}
    & \textbf{KID$\downarrow$}
    & \textbf{BKID$\downarrow$}
    & \textbf{$\Delta$CER$\downarrow$}
    & \textbf{HWD$\downarrow$} 
    & \textbf{$\Delta$I-LPIPS$\downarrow$}\\
    \midrule
    \textbf{TS-GAN}          & 42.12 & 31.97 & 43.15 & 32.44 & 0.13 & 3.07 & ~2.02 \\
    \textbf{HiGAN+}          & 78.44 & 39.47 & 80.39 & 36.50 & 0.12 & 3.07 & 53.91 \\
    \textbf{HWT}             & 31.22 & 16.73 & 26.14 & 14.44 & 0.38 & 2.59 & 10.23 \\
    \textbf{VATr}            & 34.40 & 24.64 & 32.21 & 25.01 & 0.06 & 2.36 & ~8.77 \\
    \textbf{VATr++}          & 35.53 & 19.87 & 34.15 & 16.08 & 0.12 & 2.18 & 13.30 \\
    \textbf{One-DM}          & 60.45 & 26.58 & 64.13 & 26.76 & 0.06 & 2.66 & 88.94 \\
    \textbf{DiffPen}         & 40.40 & 17.50 & 38.21 & 18.30 & \textbf{0.01} & 2.99 & 51.58 \\
    \textbf{Emuru}           & \textbf{14.39} & \textbf{10.77} & \textbf{12.34} & \textbf{10.54} & 0.13 & \textbf{1.82} & ~\textbf{0.75} \\  
    \midrule
    & \multicolumn{7}{c}{\textbf{RIMES}}\\ 
    \cmidrule{2-8}
    & \textbf{FID$\downarrow$}
    & \textbf{BFID$\downarrow$}
    & \textbf{KID$\downarrow$}
    & \textbf{BKID$\downarrow$}
    & \textbf{$\Delta$CER$\downarrow$}
    & \textbf{HWD$\downarrow$} 
    & \textbf{$\Delta$I-LPIPS$\downarrow$}\\
    \midrule
    \textbf{TS-GAN}          & 109.04 & 36.39 & 132.90 & 41.64 & 0.12 & 3.26 & 93.87 \\
    \textbf{HiGAN+}          & 160.57 & 47.38 & 183.82 & 46.23 & 0.14 & 3.39 & 48.48 \\
    \textbf{HWT}             & 118.21 & 35.26 & 128.66 & 35.60 & 0.45 & 3.36 & ~\textbf{1.89} \\
    \textbf{VATr}            & 113.76 & 30.21 & 114.21 & 27.88 & 0.07 & 3.09 & ~7.88 \\
    \textbf{VATr++}          & 110.04 & 35.61 & 104.05 & 31.90 & 0.10 & 2.83 & 26.84 \\
    \textbf{One-DM}          & 121.18 & 36.07 & 121.67 & 34.68 & 0.20 & 3.36 & 86.49 \\
    \textbf{DiffPen}         & ~89.79 & 18.25 & ~94.78 & 18.49 & \textbf{0.04} & 2.58 & 78.41 \\
    \textbf{Emuru}           & ~\textbf{26.93} & \textbf{13.36} & ~\textbf{21.19} & ~\textbf{9.40} & 0.25 & \textbf{2.18} & 47.96 \\
    \bottomrule
    \end{tabular}
     }
    \caption{Comparison on the line-level CVL and RIMES datasets between Emuru and State-of-the-Art approaches trained on IAM. KID, BKID, and $\Delta$I-LPIPS are multiplied by $10^3$ and the best performance is in bold for readability.}
    \label{tab:cvl_rimes_supp}
\end{table}
\begin{table}[]
    \centering
    \setlength{\tabcolsep}{.18em}
    \resizebox{\columnwidth}{!}{%
    \begin{tabular}{l ccccccc}
    \toprule
    & \multicolumn{7}{c}{\textbf{Karaoke Handwritten}}\\ 
    \cmidrule{2-8}
    & \textbf{FID$\downarrow$}
    & \textbf{BFID$\downarrow$}
    & \textbf{KID$\downarrow$}
    & \textbf{BKID$\downarrow$}
    & \textbf{$\Delta$CER$\downarrow$}
    & \textbf{HWD$\downarrow$} 
    & \textbf{$\Delta$I-LPIPS$\downarrow$}\\
    \midrule
    \textbf{TS-GAN}          & ~60.30 & 12.68 & ~64.80 & ~6.75 & 0.05 & 4.59 & 75.30 \\
    \textbf{HiGAN+}          & 125.75 & 69.41 & 136.75 & 72.14 & 0.04 & 4.90 & 49.49 \\
    \textbf{HWT}             & ~62.69 & 43.03 & ~59.35 & 43.98 & 0.34 & 4.50 & 33.89 \\
    \textbf{VATr}            & ~72.22 & 47.66 & ~67.70 & 46.23 & 0.04 & 3.89 & 60.46 \\
    \textbf{VATr++}          & ~67.16 & 46.53 & ~58.57 & 42.59 & 0.01 & 3.96 & 84.57 \\
    \textbf{One-DM}          & ~59.73 & 38.30 & ~56.55 & 37.93 & 0.24 & 4.31 & 47.65 \\
    \textbf{DiffPen}         & ~34.19 & 25.78 & ~28.91 & 24.03 & 0.16 & 4.18 & 33.33 \\
    \textbf{Emuru}           & ~\textbf{13.87} & ~\textbf{7.99} & ~~\textbf{9.24} & ~\textbf{5.37} & \textbf{0.13} & \textbf{2.24} & ~\textbf{0.73} \\  
    \midrule
    & \multicolumn{7}{c}{\textbf{Karaoke Typewritten}}\\ 
    \cmidrule{2-8}
    & \textbf{FID$\downarrow$}
    & \textbf{BFID$\downarrow$}
    & \textbf{KID$\downarrow$}
    & \textbf{BKID$\downarrow$}
    & \textbf{$\Delta$CER$\downarrow$}
    & \textbf{HWD$\downarrow$} 
    & \textbf{$\Delta$I-LPIPS$\downarrow$}\\
    \midrule
    \textbf{TS-GAN}          & 141.41 & 75.78 & 157.33 & 80.14 & 0.02 & 4.70 & 235.45 \\
    \textbf{HiGAN+}          & 135.34 & 63.39 & 146.34 & 65.81 & 0.03 & 5.19 & ~85.45 \\
    \textbf{HWT}             & ~72.78 & 37.40 & ~62.77 & 31.51 & 0.39 & 4.57 & 138.76 \\
    \textbf{VATr}            & ~80.38 & 41.02 & ~70.46 & 37.26 & 0.04 & 4.14 & 115.80 \\
    \textbf{VATr++}          & ~76.03 & 41.69 & ~63.17 & 36.50 & \textbf{0.01} & 4.15 & ~89.74 \\
    \textbf{One-DM}          & ~70.75 & 44.06 & ~60.90 & 42.78 & 0.25 & 4.80 & 119.63 \\
    \textbf{DiffPen}         & ~78.07 & 61.16 & ~67.17 & 61.05 & 0.14 & 4.71 & 187.68 \\
    \textbf{Emuru}           & ~~\textbf{9.85} & ~\textbf{4.33} & ~~\textbf{5.60} & ~\textbf{1.24} & 0.11 & \textbf{1.28} & ~~\textbf{5.07} \\
    \bottomrule
    \end{tabular}
     }
    \caption{Comparison on the line-level Karaoke dataset, separating calligraphy and typewritten styles, between Emuru and State-of-the-Art approaches trained on IAM. KID, BKID, and $\Delta$I-LPIPS are multiplied by $10^3$ and the best performance is in bold for readability.}
    \label{tab:karaoke_supp}
\end{table}

\section{Additional Qualitative Results}
\label{sec:qualitatives_supplementary}
\begin{figure*}
    \centering
    \begin{tabular}{cc}
         \includegraphics[width=.5\textwidth]{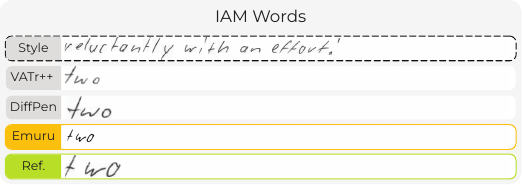} &
         \includegraphics[width=.5\textwidth]{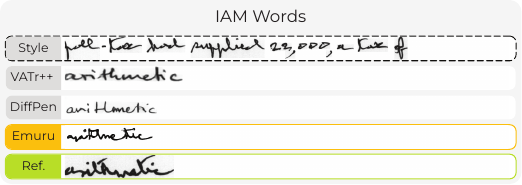} \\
         \includegraphics[width=.5\textwidth]{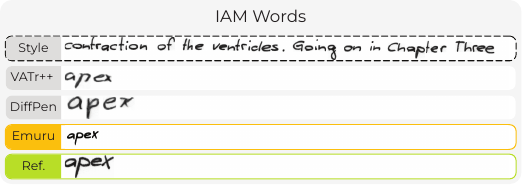} &
         \includegraphics[width=.5\textwidth]{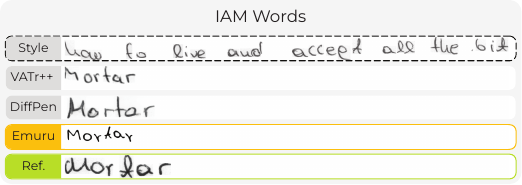} \\
         \includegraphics[width=.5\textwidth]{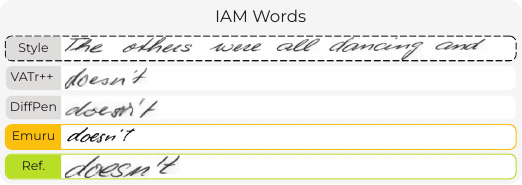} &
         \includegraphics[width=.5\textwidth]{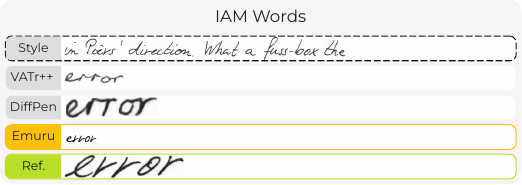} 
    \end{tabular}
    \caption{Further qualitative results on IAM words.}
    \label{fig:qualitatives_iamw}
\end{figure*}

\begin{figure*}
    \centering
    \begin{tabular}{cc}
         \includegraphics[width=.5\textwidth]{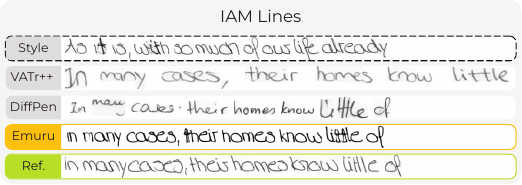} &
         \includegraphics[width=.5\textwidth]{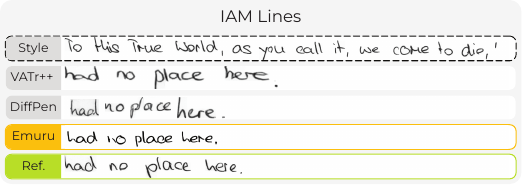} \\
         \includegraphics[width=.5\textwidth]{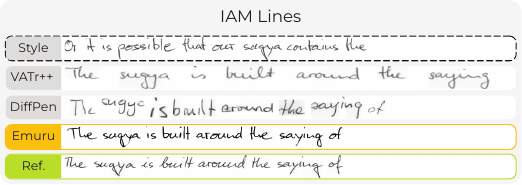} &
         \includegraphics[width=.5\textwidth]{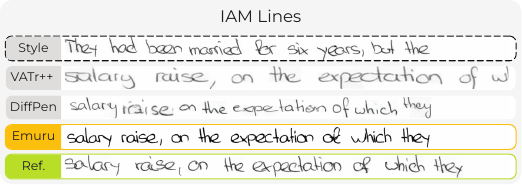} \\
         \includegraphics[width=.5\textwidth]{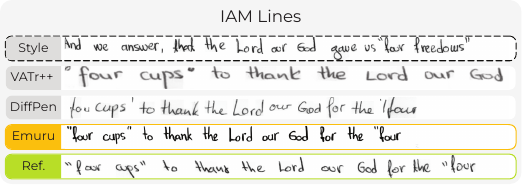} &
         \includegraphics[width=.5\textwidth]{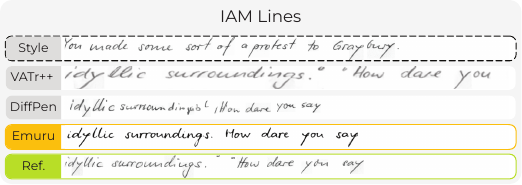} 
    \end{tabular}
    \caption{Further qualitative results on IAM lines.}
    \label{fig:qualitatives_iaml}
\end{figure*}

\begin{figure*}
    \centering
    \begin{tabular}{cc}
         \includegraphics[width=.5\textwidth]{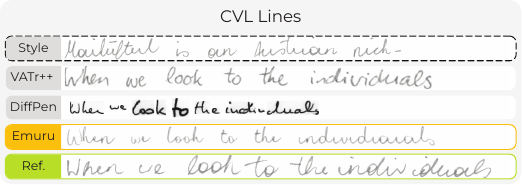} &
         \includegraphics[width=.5\textwidth]{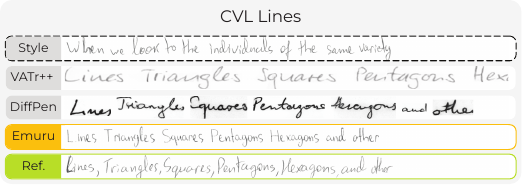} \\
         \includegraphics[width=.5\textwidth]{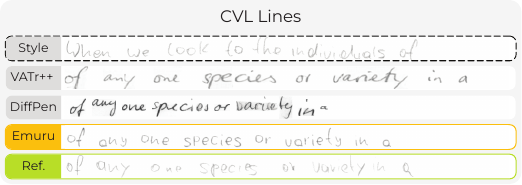} &
         \includegraphics[width=.5\textwidth]{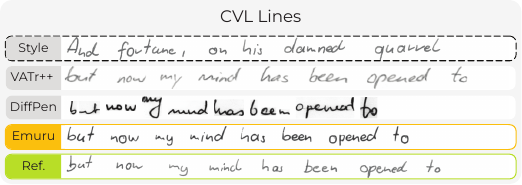} \\
         \includegraphics[width=.5\textwidth]{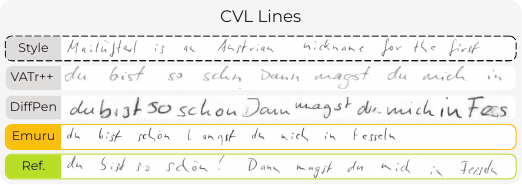} &
         \includegraphics[width=.5\textwidth]{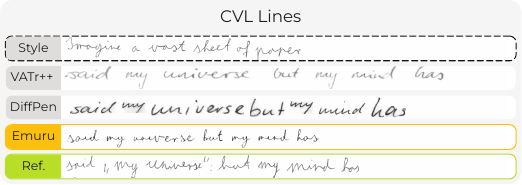} 
    \end{tabular}
    \caption{Further qualitative results on CVL lines.}
    \label{fig:qualitatives_cvl}
\end{figure*}

\begin{figure*}
    \centering
    \begin{tabular}{cc}
         \includegraphics[width=.5\textwidth]{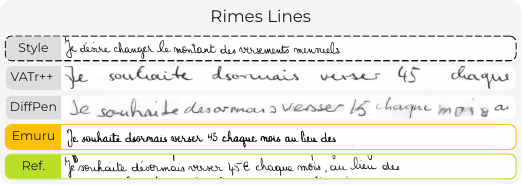} &
         \includegraphics[width=.5\textwidth]{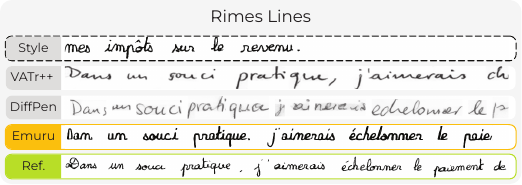} \\
         \includegraphics[width=.5\textwidth]{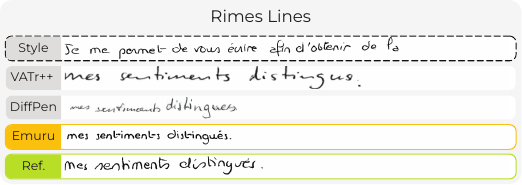} &
         \includegraphics[width=.5\textwidth]{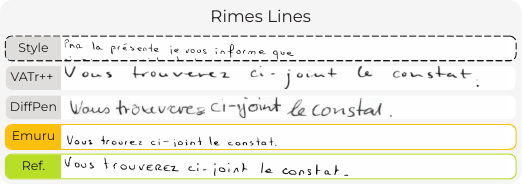} \\
         \includegraphics[width=.5\textwidth]{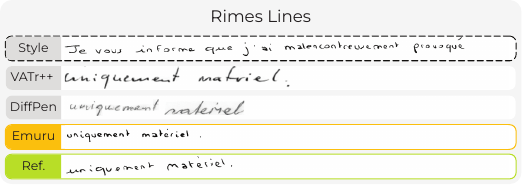} &
         \includegraphics[width=.5\textwidth]{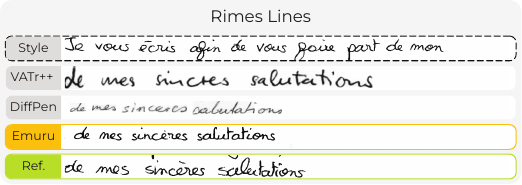} 
    \end{tabular}
    \caption{Further qualitative results on RIMES lines.}
    \label{fig:qualitatives_rimes}
\end{figure*}

\begin{figure*}
    \centering
    \begin{tabular}{cc}
         \includegraphics[width=.5\textwidth]{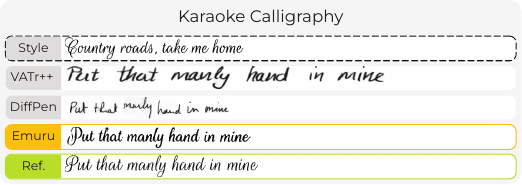} &
         \includegraphics[width=.5\textwidth]{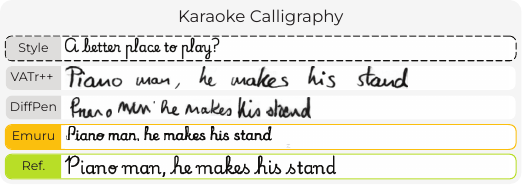} \\
         \includegraphics[width=.5\textwidth]{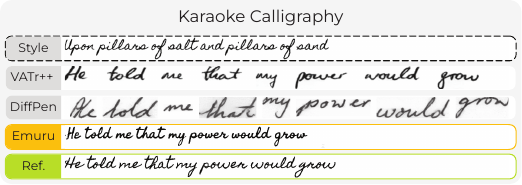} &
         \includegraphics[width=.5\textwidth]{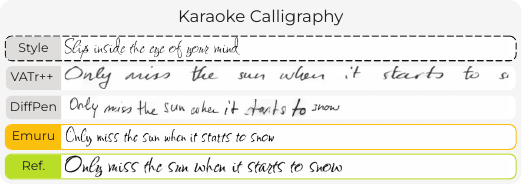} \\
         \includegraphics[width=.5\textwidth]{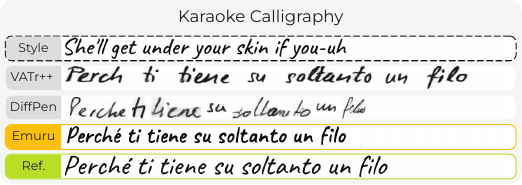} &
         \includegraphics[width=.5\textwidth]{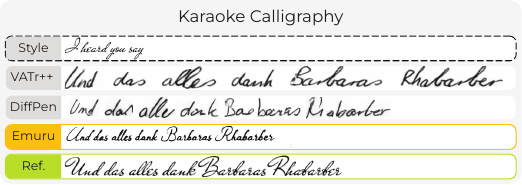} 
    \end{tabular}
    \caption{Further qualitative results on Karaoke Calligraphy.}\vspace{.85cm}
    \label{fig:qualitatives_karaokec}
\end{figure*}

\begin{figure*}
    \centering
    \begin{tabular}{cc}
         \includegraphics[width=.5\textwidth]{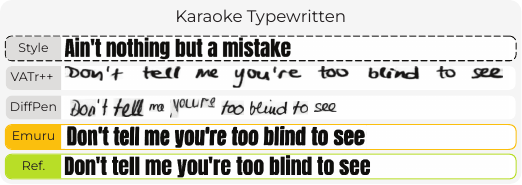} &
         \includegraphics[width=.5\textwidth]{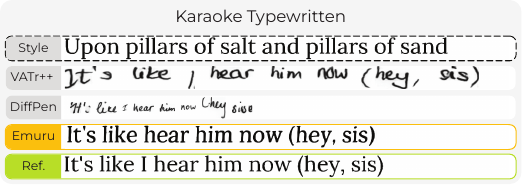} \\
         \includegraphics[width=.5\textwidth]{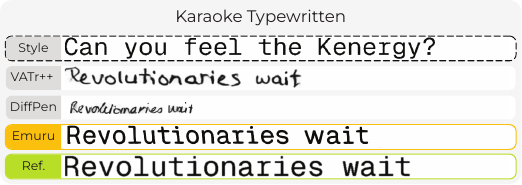} &
         \includegraphics[width=.5\textwidth]{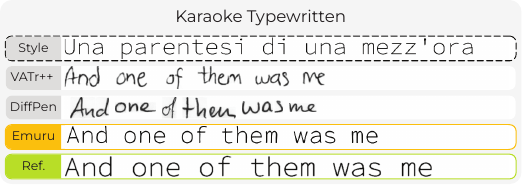} \\
         \includegraphics[width=.5\textwidth]{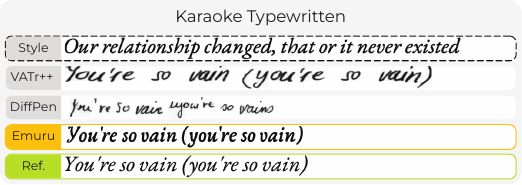} &
         \includegraphics[width=.5\textwidth]{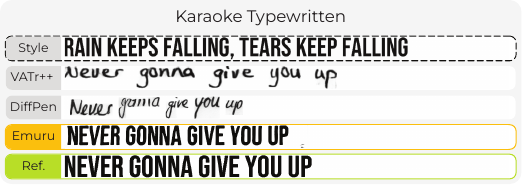} 
    \end{tabular}
    \caption{Further qualitative results on Karaoke Typewritten.}
    \label{fig:qualitatives_karaoket}
\end{figure*}

In~\Cref{fig:qualitatives_iamw,fig:qualitatives_iaml,fig:qualitatives_cvl,fig:qualitatives_rimes,fig:qualitatives_karaokec,fig:qualitatives_karaoket}, we report additional qualitative results comparing Emuru, the GAN-based VATr++~\cite{vanherle2024vatr++}, and the Diffusion Model-based DiffPen~\cite{nikolaidou2024diffusionpen} when generating images from the considered datasets. For each sample, we report the input style image used for guiding the generation and another reference image in the same style. We let the models generate the same text as in the reference to better observe the style imitation capabilities of the models.
\clearpage
\clearpage
\clearpage

\section{Editing Application}
\begin{figure}
    \centering
    \includegraphics[width=\linewidth]{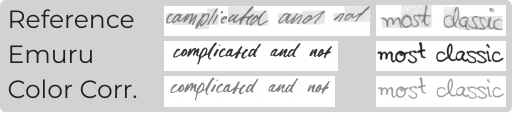}
    \caption{Thanks to the characteristics of the VAE, ink color can be easily corrected on Emuru-generated images to look more similar to the color in the reference image.}
    \label{fig:color_correction}
\end{figure}

Due to our Emuru VAE, there can be a difference in ink color between the reference image, $I_{style}$, and the Emuru-generated one, $I_{out}$. However, this difference can be easily recovered. Note that, once passed through the VAE, both $I_{style}$ and $I_{out}$ have a perfectly white background. Therefore, the background can be easily removed from both images via standard chroma keying on white. Then, the average color of the ink pixels can be computed from the reference image and directly applied to the ink pixels in the generated image (see~\Cref{fig:color_correction}).

\end{document}